\documentclass[letterpaper, 10 pt, conference]{ieeeconf}  

\IEEEoverridecommandlockouts                              

\usepackage[T1]{fontenc}
\usepackage[utf8]{inputenc}
\usepackage[hyphens]{url}
\usepackage[hidelinks]{hyperref}
\hypersetup{breaklinks=true}

\newcommand{\myvec}{\vec{}}

\usepackage{stmaryrd}

\usepackage{tikz}
\usepackage{subcaption}
\usepackage{graphicx}
\usepackage{dsfont}
\usepackage{eucal}

\newcommand{\minisection}[1]{\textbf{#1}\hspace{0.3em}}
\newcommand{\mlp}{\textit{MLP}}
\newcommand{\renn}{\textit{ReNN}}

\usepackage{amsmath,amsfonts,bm}

\def\eqref#1{equation~\ref{#1}}

\def\1{\bm{1}}

\DeclareMathAlphabet{\mathsfit}{\encodingdefault}{\sfdefault}{m}{sl}
\SetMathAlphabet{\mathsfit}{bold}{\encodingdefault}{\sfdefault}{bx}{n}

\title{\LARGE \bf
Towards Practical Multi-Object Manipulation using Relational Reinforcement Learning
}

\author{Richard Li$^{1}$ and Allan Jabri$^{2}$ and Trevor Darrell$^{2}$ and Pulkit Agrawal$^{1}$
\thanks{$^{1}$Richard Li and Pulkit Agrawal are at the Massachusetts Institute of Technology, USA,
{\tt\small \{rli14, pulkitag\}@mit.edu}}%
\thanks{$^{2}$Allan Jabri and Trevor Darrell are at the University of California Berkeley,
        {\tt\small \{ajabri, trevor\}@berkeley.edu }}%
}

\begin{document}

\let\oldtwocolumn\twocolumn
\renewcommand\twocolumn[1][]{%
    \oldtwocolumn[{#1}{
    \begin{flushleft}
           \includegraphics[width=0.98\textwidth]{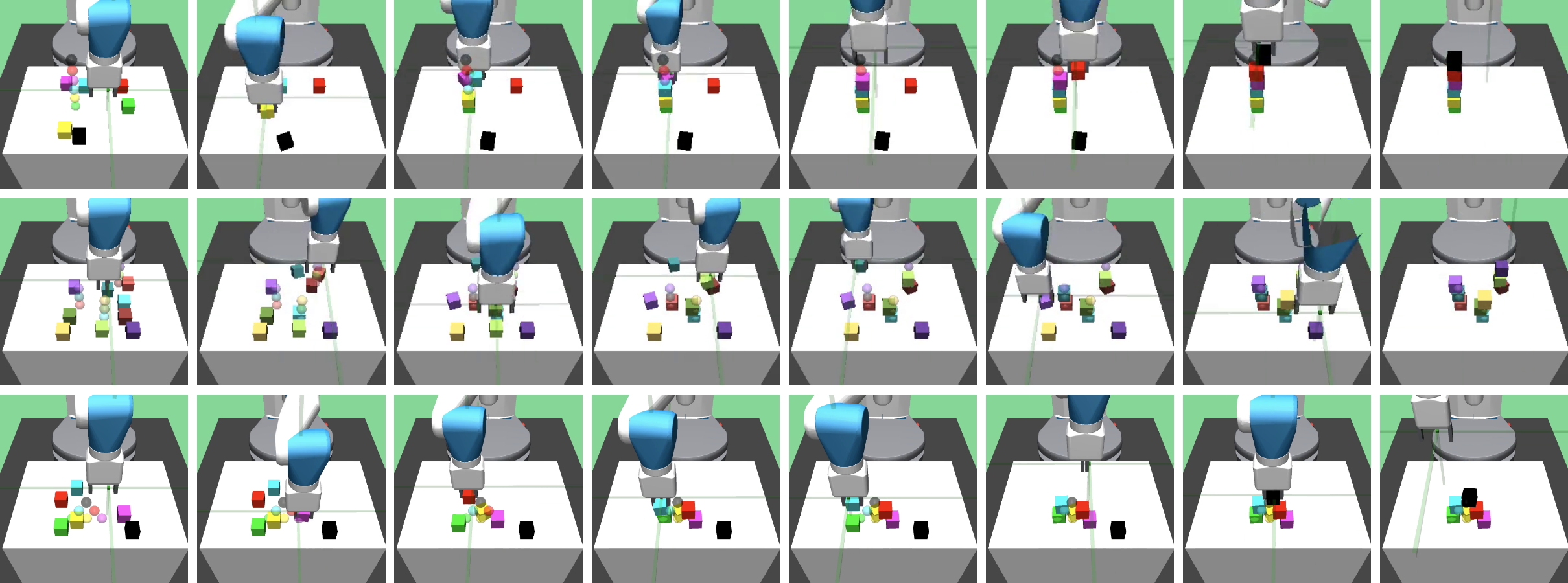}
           \captionof{figure}{We present a reinforcement learning system that can stack 6 blocks without requiring any demonstrations or task-specific assumptions. The last two rows show zero-shot generalization results of configuring blocks into unseen configurations of multiple towers and pyramids without additional training. See the videos here: \url{https://richardrl.github.io/relational-rl}. 
           }
           \label{fig:fig1}
            \end{flushleft}
    }]
}

\maketitle
\thispagestyle{empty}
\pagestyle{empty}

\begin{abstract}
Learning robotic manipulation tasks using reinforcement learning with sparse rewards is currently impractical due to the outrageous data requirements. Many practical tasks require manipulation of multiple objects, and the complexity of such tasks increases with the number of objects. Learning from a curriculum of increasingly complex tasks appears to be a natural solution, but unfortunately, does not work for many scenarios. We hypothesize that the inability of the state-of-the-art algorithms to effectively utilize a task curriculum stems from the absence of inductive biases for transferring knowledge from simpler to complex tasks. We show that graph-based relational architectures overcome this limitation and enable learning of complex tasks when provided with a simple curriculum of tasks with increasing numbers of objects. We demonstrate the utility of our framework on a simulated block stacking task. Starting from scratch, our agent learns to stack six blocks into a tower. Despite using step-wise sparse rewards, our method is orders of magnitude more data-efficient and outperforms the existing state-of-the-art method that utilizes human demonstrations. Furthermore, the learned policy exhibits zero-shot generalization, successfully stacking blocks into taller towers and previously unseen configurations such as pyramids, without any further training.
\end{abstract}

\section{INTRODUCTION}
The main idea in reinforcement learning is to incentivize actions that maximize rewards. Unlike video games, where rewards are readily available, for manipulation tasks, a reward function must be manually constructed. For example, to pick and place a block, the rewards might be inversely proportional to the manipulator's distance from the block and the block's distance from the target location. Such rewards that frequently provide information about the task are known as \textit{dense rewards}. Using dense rewards, an agent can get stuck in a local minimum and never complete the desired task. It is well known that intuitively reasonable reward functions can often result in unexpected or undesirable behaviors~\cite{hadfield2017inverse}. This makes reward design very challenging. 

An alternative is to provide the agent with \textit{sparse rewards}, which may either be provided after the agent completes the overall task (i.e., terminal reward) or extremely intermittently when the agent completes critical steps (i.e., step-wise rewards). \textit{Sparse rewards} are more straightforward to define than dense rewards. However, because many tasks require execution of a long sequence of actions, sparse rewards drastically complicate the challenges of exploration and credit-assignment. Training with \textit{sparse rewards}, therefore, either completely fails or requires massive amounts of data. 

Practical reinforcement learning systems have sidestepped the challenge of learning from sparse rewards by either using (a) human demonstrations, (b) sim2real transfer, (c) careful environmental instrumentation to simplify the task or (d) meticulous reward shaping. Using these ideas, RL has been applied to wide variety of robotic tasks such as stacking blocks in a tower~\cite{deisenroth2011learning,rusu2016progressive,nair2017overcoming}, opening doors~\cite{gu2017deep}, flipping pan-cakes~\cite{kormushev2010robot}, hitting a ball, orienting a cube~\cite{andrychowicz2018learning} and other dexterous manipulation tasks~\cite{rajeswaran2017learning}.

Developing data-efficient algorithms that can learn from sparse rewards will alleviate the need for demonstrations and painful reward design. It will consequently open up many application areas, where RL cannot be applied today. Past works have improved learning efficiency of RL algorithms using better optimization methods~\cite{hessel2018rainbow,sac,schulman2017proximal}, combining model-based and model-free learning~\cite{levineFDA15}, hierarchical learning~\cite{suttonoptions1999}, and design of better exploration methods~\cite{schmidhuber2010formal,lopes2012exploration,oudeyer2009intrinsic,pathak2017curiosity,sukhbaatar2018intrinsic}. A few recent works used the compositional task structure to improve the data efficiency of RL algorithms~\cite{zambaldi2019deep,wang2018nervenet,pathak19assemblies}. 

In the related field of supervised deep learning, transfer of knowledge by pre-training on a source task followed by finetuning on a target task~\cite{donahue2014decaf,agrawal2016learning,ren2015faster} has been very successful in reducing the data requirements.  However, in the context of RL, learning from multiple tasks and transferring this knowledge to reduce data requirements for a new task remains an open challenge~\cite{zhang2018study,cobbe2018quantifying,justesen2018illuminating,nichol2018first}. One potential reason for lack of transfer is that learning from a new task exacerbates the already existing problem of credit assignment. The inability to assign credit, in turn, increases the variance in the gradients and consequently results in learning failure~\cite{ghosh2017divide}. One solution is to pace the agent's learning, where it only gets a new task when it has mastered previous tasks (i.e., curriculum learning~\cite{bengio2009curriculum}). 

It turns out that in RL settings, curriculum learning is also not straightforward. To better understand why it is the case, consider solving the problem of stacking multiple blocks into a tower using a curriculum of stacking an increasing number of blocks.  Suppose the agent has mastered the skill of stacking two blocks. The introduction of the third block preserves the task structure but changes the distribution of the agent's input. In the absence of appropriate inductive biases to deal with changes in inputs, the agent resorts to treating the new data distribution as a new learning problem and is unable to leverage its knowledge from past tasks efficiently.

One well-known method to tackle changing data distribution is training with data-augmentation. In RL settings, this idea has translated into domain randomization~\cite{tobin2017domain}. In the running example, training with randomization involves sampling a random number of blocks from a uniform distribution in every episode. However, because, in most episodes, the agent would be tasked to stack multiple blocks, learning in such a situation remains very challenging. This consideration suggests that the major hindrance in learning from a curriculum may not be in the design of the curriculum, but the inability of learning systems to transfer knowledge across the different tasks in the curriculum. 

In this work, we show that training a policy represented by a attention based graph neural network (GNN) overcomes the challenges associated with curriculum learning in multi-object manipulation tasks. Our agent learns to stack six or more blocks from scratch (see Figure~\ref{fig:fig1}). We use a simple curriculum strategy, which increases the number of blocks when the agent masters a target task with a fewer number of blocks. The attention-based GNN complements the curriculum by providing the appropriate inductive bias to transfer knowledge between tasks with a different number of objects. To the best of our knowledge, ours is the first work to solve the problem of stacking six or more blocks using RL and without requiring any expert demonstrations. Our method is orders of magnitude more efficient than the previous state-of-the-art method relying on human-provided demonstrations~\cite{nair2017overcoming}.

Furthermore, our system can build towers that are taller than the training time. It also succeeds at placing blocks in different configurations such as pyramids without any additional training (i.e., \textit{zero-shot generalization}). While we present results on the task of stacking blocks in various arrangements, the approach developed in this work does not make any task-specific assumption and is therefore applicable to a wide range of tasks involving manipulation of multiple objects. 

\section{Related Work}
\label{sec:related_works}
Our work is broadly related to techniques for scaling reinforcement learning algorithms to more complex robotic manipulation settings, as well as the use of relational and curricular inductive biases in machine learning.

\noindent \minisection{Relational Inductive Bias:} The use of relational inductive biases has a long history in reinforcement learning \cite{kaelbling2001learning, dzeroski2001, van2002relational}, and more broadly in logic and machine learning \cite{Russell_Norvig_2009}. Recently, there has been great interest in the use of Graph Neural Networks (GNNs) for representing graph data structures, which are especially suitable for object-oriented environments~\cite{bruna14, defferrard16, kipf2016semi,structuralrnn16,transformer, battaglia16}. In the context of RL, a key motivation for relational representation is to support a varying number of objects as inputs and to explicitly model relationships between objects. In the past, GNNs have been studied in context of learning and transferring policies for locomotion across agents with variable morphologies~\cite{wang2018nervenet,pathak19assemblies}. 

Closest to our work is past research combining GNNs with policy learning for manipulation tasks. However these works either rely on tens or hundreds of thousands of expert demonstrations \cite{oneshotimitationlearning, janner2019reasoning} or exclusively show results on video games\cite{zambaldi2019deep}. Furthermore, while these works have considered GNNs to improve efficiency of solving a single task, we combine GNNs with learning from a curriculum of increasingly complex tasks to solve long-horizon manipulation problems that cannot be solved directly using current methods. 
 
\noindent \minisection{Curriculum Learning:} 
Curriculum learning addresses the effect of data sampling strategies on learning, under the presumption that proper sampling of tasks can allow for more sample efficient learning and avoidance of local minima \cite{elman1993}. In particular, prior work has shown that ordering tasks by heuristic measures of difficulty can be effective \cite{Bengio2009, zaremba14}. A line of work has studied automatic discovery of curricula based on learning progress \cite{graves17}, adversarial self-play \cite{sukhbaatar2018intrinsic, goalgan17}, or backtracking \cite{florensa2017reverse}. So far, these methods have not yielded curricula capable of automatically discovering tasks of the complexity we consider. In this paper, our contribution is not in proposing a new algorithm or heuristic for choosing the task curricula, but to demonstrate the graph-based representations can make use of a curriculum for learning complex tasks. 

\noindent \minisection{Block Stacking:} 
Prior work on block stacking either heavily relied on human demonstrations~\cite{nair2017overcoming,duan2017one}, or required significant reward engineering~\cite{rusu2016progressive,popov2018dataefficient}, and/or carefully designed curriculum~\cite{rusu2016progressive} of reaching, picking and placing blocks. Such design of curriculum and reward functions are hard problems with no known principled solutions. The work of ~\cite{deisenroth2011learning} stacked blocks using a low-cost robot. However, they assumed the blocks were already picked and used a dense reward function. Other lines of work~\cite{kroemer2018kernel,toussaint2015logic} achieved impressive results on stacking objects, but relied on extensive human-defined knowledge of detecting keypoints or assuming access to physics simulation. In contrast, we present a simple but effective method for stacking blocks using RL that makes minimal assumptions about task structure or the environment.  

\noindent \minisection{Hierarchical Reinforcement Learning (HRL)}
aims to address the scaling and generalization problem in RL by decomposing problems into smaller subproblems. Examples of HRL frameworks include the ``options" framework \cite{suttonoptions1999}, feudal learning~\cite{feudalnets,dayan1993feudal} and the MaxQ framework \cite{maxq}. 
A key unsolved challenge is joint end-to-end learning of multiple levels of control, while avoiding degenerate solutions that lack hierarchical abstraction. Most successful instantiations of hierarchical RL make use of domain knowledge to construct a hierarchy~\cite{bacon16}. To our knowledge, no HRL algorithms have been successful at stacking tasks of the complexity we consider \cite{modulatedpolicyhierarchies}. 

\section{Experimental Setup}
\label{sec: experimental_setup}
Figure~\ref{fig:fig1} shows our simulated robotic environment consisting of a 7-DoF Fetch robot arm equipped with a two-fingered parallel jaw gripper based on OpenAI's $\tt{FetchPickAndPlace}$~\cite{plappert2018multi}. MuJoCo physics engine~\cite{mujoco} was used for simulations. The robot is tasked to manipulate 1-9 blocks kept on a table. Each block is a cube with sides of 5cm. The robot's action space is 4D, consisting of relative change in 3D position of its end-effector and a scalar value representing the distance between two fingers of the gripper. 

\noindent \minisection{Observations:} The agent observes gripper features $X^{ee}$, including gripper velocity and position, and features representing N blocks. The block features are denoted by $X^f:{x^f_1, x^f_2, .. x^f_N}$, where $N \in [1, 9]$ and $x^f_i$ is the feature representation of the $i^{th}$ block. Each block is represented by a 15-D vector consisting of 3D position $(x^p_i)$, 3D orientation expressed as Euler angles, 3D position relative to the gripper, 3D cartesian velocity and 3D angular velocity. The goal is expressed as set of 3D block positions, $X^g: {x^g_1, x^g_2, .. x^g_N}$. The overall input to the agent is therefore $\{X^{ee}, X^f, X^g\}$. At the start of every episode, the initial block positions are randomly initialized on the table and the goal positions are sampled using a pre-determined distribution. The maximum length of every episode is $50*N$ steps, where $N$ is the number of blocks. 

\noindent \minisection{Reward:} We use a step-wise sparse reward function where the robot is only rewarded when it places the $i^{th}$ block within a distance of $\delta$ from its desired goal location. The overall reward for placing $N$ blocks is given by: $\sum_i \mathds{1}_{\lVert x^p_i - x^g_i \rVert <\delta}$. We noticed that with this reward function, the robot learns to hold the top two blocks in its gripper instead of placing them and moving its hand away. To discourage this behavior, we added an additional term $\mathds{1}_{\rm grip\_away}$ in the reward function to encourage the robot to move its hand away from the tower. This additional penalty was only provided when the hand was at a distance greater than $2\delta$ from a ``fully-stacked" tower. The overall reward is therefore given by,     
    $r_t =  \sum_i \mathds{1}_{\lVert x_i - g_i \rVert <\delta} - \mathds{1}_{\rm grip\_away}$. 
Following~\cite{plappert2018multi}, we set $\delta$ = 5cm, the size of each block. 

\section{Preliminaries}
\label{sec:background}

\subsection{Reinforcement Learning}
A typical RL agent acts within an environment \textit{E}, modeled by a discrete-time Markov Decision Process (MDP) described by state space $\mathcal{S}$, action space $\mathcal{A}$, transition function $\mathcal{T}$, reward function ${r(s,a)}$, and discounting factor ${\gamma}$. The aim of the agent is to maximize the expected cumulative reward along states $s_{1:T}$ caused by a sequence of actions $a_{1:T-1}$, by learning a suitable policy $a_t = \pi(s_t)$, i.e.
$ \max_\pi \mathbb{E}_{a \sim \pi, s \sim \mathcal{T}} [\sum_{t=1}^T \gamma^{(t-1)} r(s_t, a_t)]$.

A relatively efficient class of policy search algorithms is off-policy reinforcement learning. Q-learning~\cite{watkins1992q} is a well known choice for off-policy learning, wherein the aim is to model the Q-function, i.e. $Q(s_t,a_t) = r(s_t,a_t) + \sum_{i=t+1}^T{\gamma^{t+1-i}r(s_i, a_i)}$. In principle, the optimal Q-function is found by solving the Bellman equation \cite{sutton1998}.
In practice, we approximate the Q-function with a function approximator (i.e. a neural network) parameterized by $\theta$ by minimizing the Bellman error $\mathcal{E}(\theta) = \frac{1}{2} \lVert Q_\theta(s_{t+1}, a_{t+1}) - (r_{t} + \gamma \max_{a_t} Q_{\theta_c} (s_t, a_t)) \rVert^2$, where $\theta_c$ is an optimization constant that represents the weights of a slowly-updated "target" network.
 
\subsection{Goal-Conditioned RL}
While the above formulation is appropriate for a single goal, for solving multiple tasks, it is necessary to provide a task description as input~\cite{schaul2015universal,agrawal2016learning,her}. Goal conditioned policies are expressed as $a_t = \pi(s_t, s_g)$, where $s_g$ represents the goal state. The learning problem is expressed as:
\begin{equation}
\label{eq:multigoalobjective}
\max_\pi \mathbb{E}_{s_g\sim \rho(s_g), a \sim \pi, s \sim \mathcal{T}} [\sum_{i=t}^T \gamma^{(t-i)} r(s_t, a_t, s_g)]
\end{equation} 
where goal $s_g$ is sampled from a goal distribution $\rho(s_g)$.

\subsection{Graph Neural Networks (GNN)}
\label{subsec:attention_graph_bg}
The central computation in a GNN is message passing between 1-hop vertices of a graph, performed by a \emph{graph-to-graph} module.  This module takes as input a variable-size vertex set $\textbf{v} = \{ \myvec{v_i} \}_{i=1}^{N_v}$ and outputs an updated set $\textbf{v}' = \{ \myvec{v'_i} \}_{i=1}^{N_v}$, where $N^v$ is the number of vertices in the input graph. $\myvec{v_i}, \myvec{v'_i}$ denote feature vectors of the $i^{th}$ node before and after a round of message passing. In each message passing round, each vertex sends a message to every other vertex. In attention-based GNNs, the incoming messages are weighted by a scalar coefficient (computed by attention) according to their relevance to the receiving vertex. The new feature representation of the vertex is the weighted sum of incoming messages. Message passing is typically performed multiple times. After message passing, the entire graph is represented as a fixed-sized embedding by pooling features across all vertices.

Mathematically, let the feature representation of the $i^{th}$ vertex at timestep $t$ be $\myvec{v_i}^t$. In every message passing round, each vertex generates a query $\myvec{q_i^t}$, key $\myvec{k_i^t}$ and a message $\myvec{m_i^t}$ using independently-parameterized functions $\myvec{q_i^t} = \phi_q^t(\myvec{v_i^t})$, $\myvec{k_i^t} = \phi_k^t(\myvec{v_i})$, and $\myvec{m_i^t} = \phi_m^t(\myvec{v_i^t})$. Each vertex in the graph receives a message from all the vertices and computes it's feature representation, $\myvec{v_i^{t+1}} = \sum_j w_{ij} \myvec{m_j^t}$, where $w_{ij}$ are the attention weights and are computed as follows: $w_{ij} = \textrm{softmax}\Bigg(V^{T} \textrm{tanh}(\myvec{q_i} + \myvec{k_j})\Bigg)$.

\section{Method}
\label{sec:method}
We present a simple, but effective method for solving long-horizon, sparse reward tasks using reinforcement learning. Our core contribution is to equip the RL agent with inductive biases of relational reasoning in order to enable learning from a curriculum of tasks of increasing complexity. We use Soft-Actor Critic (SAC;~\cite{sac}) as our base learning algorithm because it is more robust to choice of hyperparameters and random seeds as compared to alternative off-policy learners such as DDPG~\cite{lillicrap2015continuous}. To use the same policy for multiple tasks, we modified SAC to be goal-conditioned~\cite{schaul2015universal,agrawal2016learning,her}. For better sample efficiency, we also incorporated the idea of goal re-labelling via hindsight experience replay (HER;~\cite{her}). Details of SAC and HER can be found in the respective papers and are not directly relevant to our work. While we use SAC + HER for policy learning, our contributions are not specific to these algorithms and are applicable to any policy learning method.

We represent both the actor and critic in SAC using the graph neural network architecture described in Section~\ref{subsec:attention_graph_bg}. The various components of the GNN ($\phi_q^t, \phi_m^t, \phi_k^t$) use 64D linear layers. We use separate weights for each round of message passing and terminate the message passing after 3 rounds. We use a residual connection and layer normalization between the output of message passing round $t$ and the input of message passing round $t+1$ to ease optimization. We call this agent architecture \renn{}. We compare the performance of \renn{} against the baseline system that constructs the actor and critic using four layers of 256D fully connected layers (referred to as \mlp{} in rest of the paper).  

\begin{figure}[t!]
\centering
\begin{tikzpicture}
  \node (img)  {\includegraphics[scale=.23]{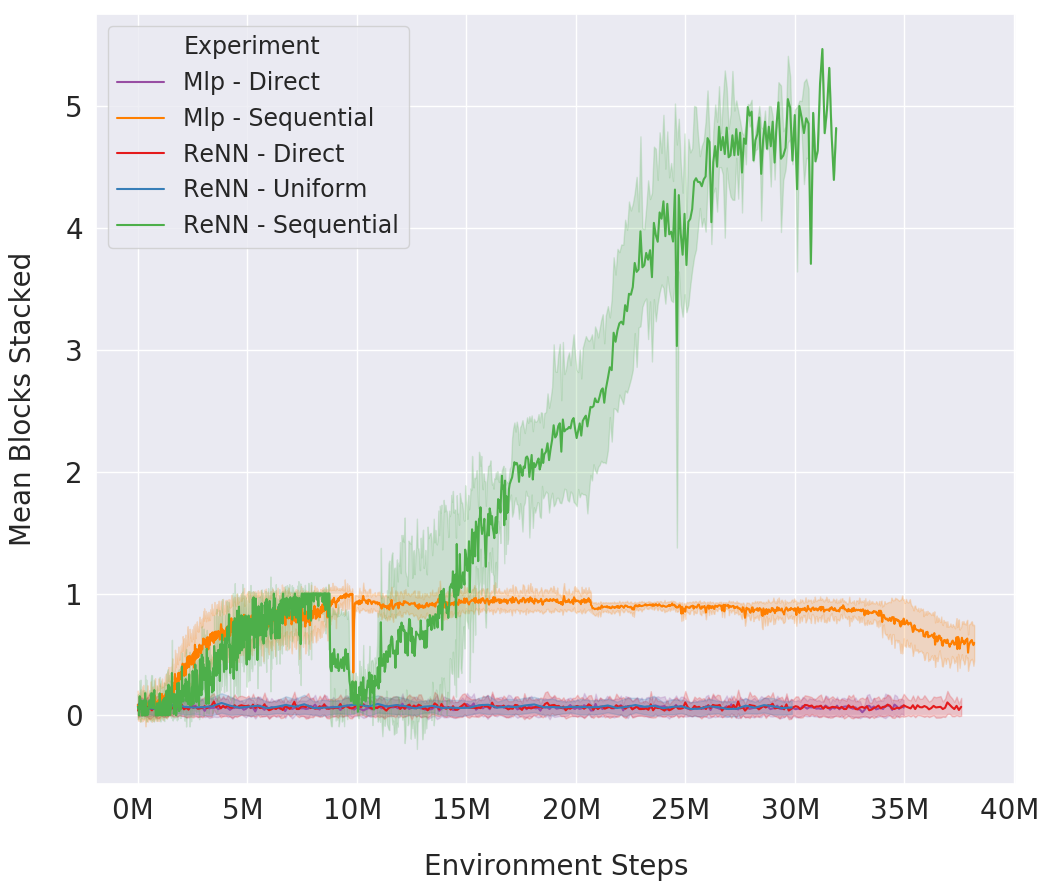}};
 \end{tikzpicture}
      \caption{Comparing task performance measured as the mean number of blocks stacked per timestep during training. We report the mean and standard deviation across multiple workers and/or seeds. The performance of relational (\renn{}) and the usual multi-layer (\mlp{}) architectures are reported when they are subjected to different training curricula described in Section~\ref{sec:training_curricula}. Both \renn{} and \mlp{} fail to stack blocks, when the robot was directly trained for stacking 6 blocks ($\mlp - Direct$, $\renn{} - Direct$). Only \renn{} trained with sequential curriculum ($\renn{} - Sequential$) succeeds at stacking six blocks.}  
    \label{fig:blocks_stacked}
    \vspace{-4mm}
\end{figure}

\noindent \minisection{Training Curriculum:} We trained the robot to stack multiple blocks using three different curricula of tasks: 
\label{sec:training_curricula}
\begin{itemize}
    \item {\textbf{Direct}: The robot was directly tasked to learn a policy to stack six blocks starting from scratch.}
    \item {\textbf{Uniform:} At every episode, the number of blocks was uniformly sampled between 1 and 6.}
    \item{\textbf{Sequential}: The robot was tasked to first pick and place a single block at goal positions that were uniformly and randomly chosen to be on the table or in the air. The robot then had to pick and place \textit{2} blocks, where goal position of one block was sampled on the table and the goal position for the second block was sampled using the process described above. Thereafter, the robot was tasked with stacking blocks in a single tower configuration starting with 2 blocks. After the robot perfected stacking (N-1) blocks, it was given N blocks to stack. N was sequentially increased from 3 to 6. The transition points in this curriculum were manually chosen based on the success rates on stacking.}
\end{itemize}

\subsection{Testing Details}
\label{sec:testing_details}
\begin{figure*}[t!]
\centering
\begin{centering}
\hspace*{.16in}
    \begin{subfigure}[t]{0.31\linewidth}
        \includegraphics[width=\linewidth]{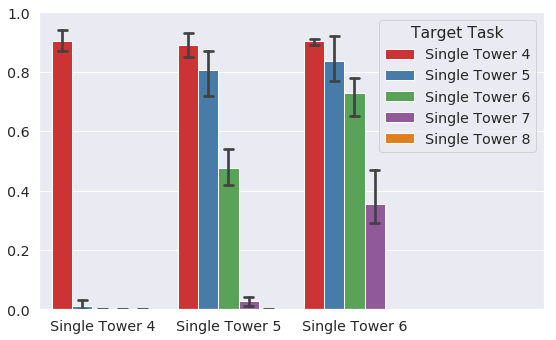}
        \caption{}
    \end{subfigure}
    \begin{subfigure}[t]{0.31\linewidth}
        \includegraphics[width=\linewidth]{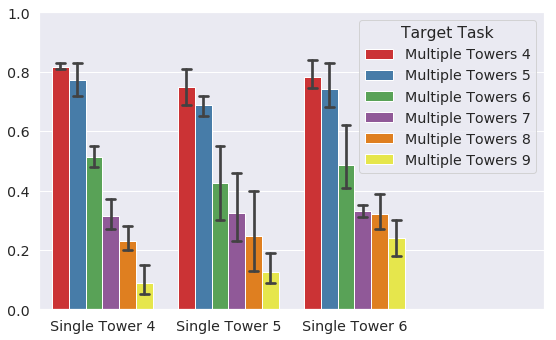}
        \caption{}
    \end{subfigure}
    \begin{subfigure}[t]{0.31\linewidth}
        \includegraphics[width=\linewidth]{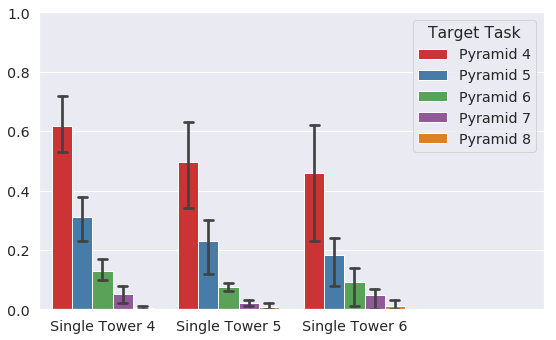}
        \caption{}
    \end{subfigure}
    \end{centering}
    \caption{Quantitative evaluation of zero-shot generalization results of policies trained to stack $i$ blocks in a single tower. These policies were evaluated, without any further training, on (a) single (but taller) tower (shown in shades of red); (b) multiple towers; (c) pyramid configurations. Details of these testing setups can be found in Section~\ref{sec:testing_details}. The results show that robot is capable of zero-shot generalization to many of these tasks.}
    \label{fig:zero_shot}
    \vspace{-4mm}
\end{figure*}

We evaluated the generalization of the policy trained for stacking a single tower by evaluating its performance on the following tests (see Appendix for visuals):

\begin{itemize}
    \item \textbf{Single Tower:} A single point was uniformly sampled on the table to serve as the base of a block tower. The goal positions of the blocks corresponded to translation along the z-axis from the base. 
    
    \item \textbf{Multiple Towers:} Few points $(k \in \{2,3\})$ were sampled on the table to serve as the base location of multiple towers. Each block was randomly assigned to a tower to produce towers of approximately equal height. 
    
    \item \textbf{Pyramid:} A uniformly sampled point on the table served as a corner point for pyramid configuration. Figure \ref{fig:goal_configs} shows different Multiple Towers and Pyramid goal configurations for varying number of blocks.  
\end{itemize}

We report performance of \renn - \textit{Sequential} (referred to as \renn ~in later text) across three seeds. For other methods we report performance on a single seed.  
Success rate is reported as accuracy of completing a task averaged over 100 episodes. An episode is counted as successful when each block is within its goal position at the final time step.

\section{Results}
\begin{table}[h!]
\caption{Comparing the performance of our method against the previous state-of-the art~\cite{nair2017overcoming} that makes use of human demonstrations on the block stacking task. Each entry, $p\%\:(s)$, denotes accuracy of $p\%$ after $s$ number of environment steps. Our method is both more sample efficient and outperforms prior work.}
\label{singletower_success_rates}
\begin{center}

\setlength\tabcolsep{4pt} 
\begin{tabular}{|c||c|c|c|c|}
\hline
Task & Single Tower 4 & Single Tower 5 & Single Tower 6\\\hline
Nair`17~\cite{nair2017overcoming} & 91\% (850M)  & 50\% (1000M) & 32\% (2300M)\\
\hline
Ours &\textbf{93\%}$\pm$4\% (\textbf{23M}) & \textbf{84\%}$\pm$6\% (\textbf{27M})&\textbf{75\%}$\pm$4\% (\textbf{30M})\\
\hline
\end{tabular}
\end{center}
\vspace{-4mm}
\label{table:results}
\end{table}

Figure~\ref{fig:blocks_stacked} shows that \renn{} trained with the sequential curriculum (green line; section \ref{sec:method}) succeeds at stacking six blocks into a tower. Standard \mlp{} architectures or \renn{} trained to directly stack 6 blocks without the curriculum fail. Our experiments revealed that training with \textit{uniform} curriculum was also insufficient. These results show that both \renn{} and the sequential training are critical for success. To the best of our knowledge, ours is the first paper to show that is is possible to train a RL agent to stack six or more blocks in a tower after starting from scratch, without requiring expert demonstrations. 

We report quantitative performance of our method and baselines in Table~\ref{table:results}. Our method achieves a success rate of 75\% at stacking 6 blocks in 30 million timesteps. In comparison, the existing state-of-art method~\cite{nair2017overcoming}, that makes use of human demonstrations and resets, achieves only a success rate of 32\% after over 2.3 billion timesteps. While the base learning algorithm used by~\cite{nair2017overcoming} is DDPG + HER, in comparison to SAC + HER used by us, the orders of magnitude difference in performance cannot be attributed to the choice of using SAC instead of DDPG. We attempted to replicate results of~\cite{nair2017overcoming} using SAC. However, we were unsuccessful at training SAC with behavior cloning due to the challenge in weighing the entropy term in SAC against the behavior cloning loss. 

Careful analysis of Figure~\ref{fig:blocks_stacked} reveals that there are several dips in performance as the training progresses. Many of the significant dips correspond to increase in task complexity to stack $\textit{N+1}$ blocks, after stacking $\textit{N}$ blocks. In most cases, the dip in performance is overcome after little additional experience. The only notable exception is the performance dip at 9M steps that corresponds to transitioning from 1 to 2 blocks. This was the first time the agent observed multiple objects. 
Additionally, We found that SAC converged faster, albeit with higher variance when it's exploration was augmented to take a random action with probability of 0.1. 

\subsection{Zero-shot Generalization}
\label{subsec:zero_shot}

\begin{figure*}[t!]
\begin{centering}
\hspace*{-.3in}
\begin{subfigure}[b]{0.124\linewidth}
    \includegraphics[width=\linewidth]{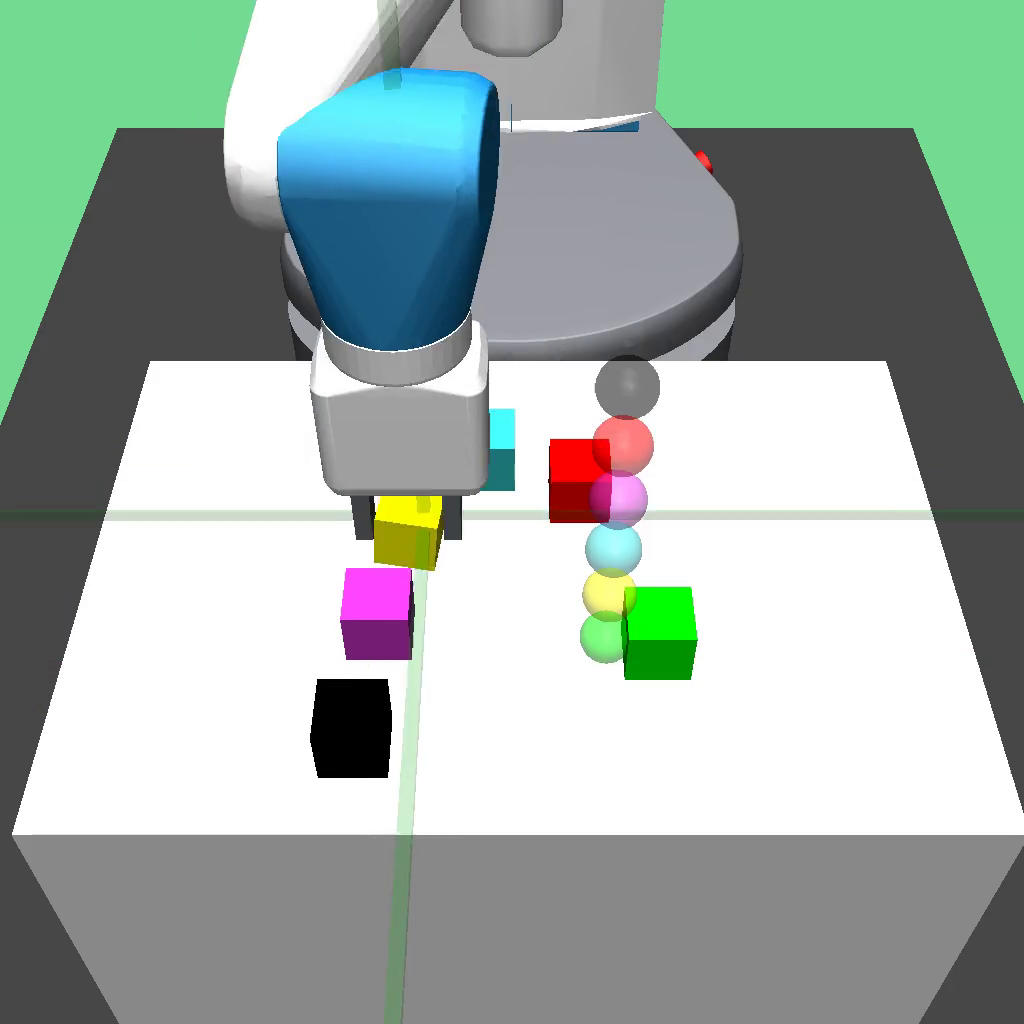}
\end{subfigure}
\begin{subfigure}[b]{0.124\linewidth}
    \includegraphics[width=\linewidth]{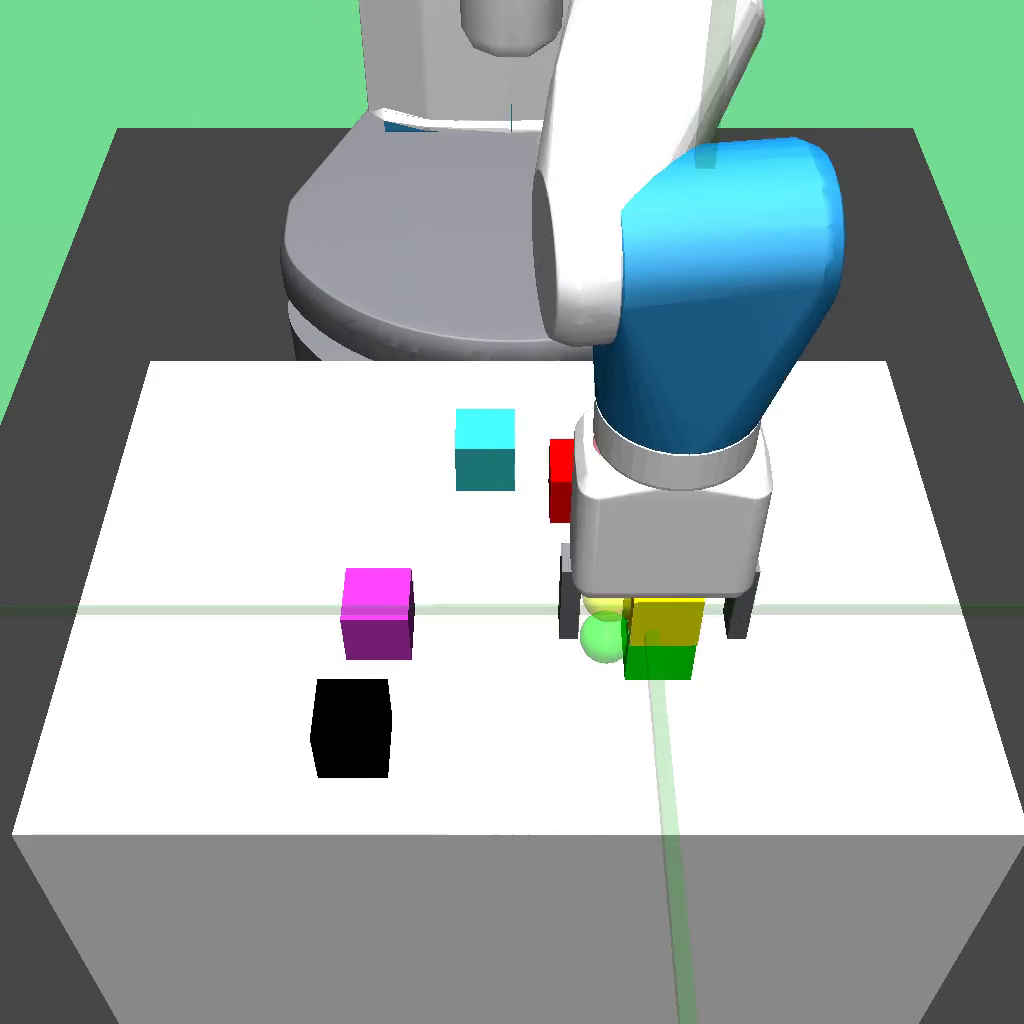}
\end{subfigure}
\begin{subfigure}[b]{0.124\linewidth}
    \includegraphics[width=\linewidth]{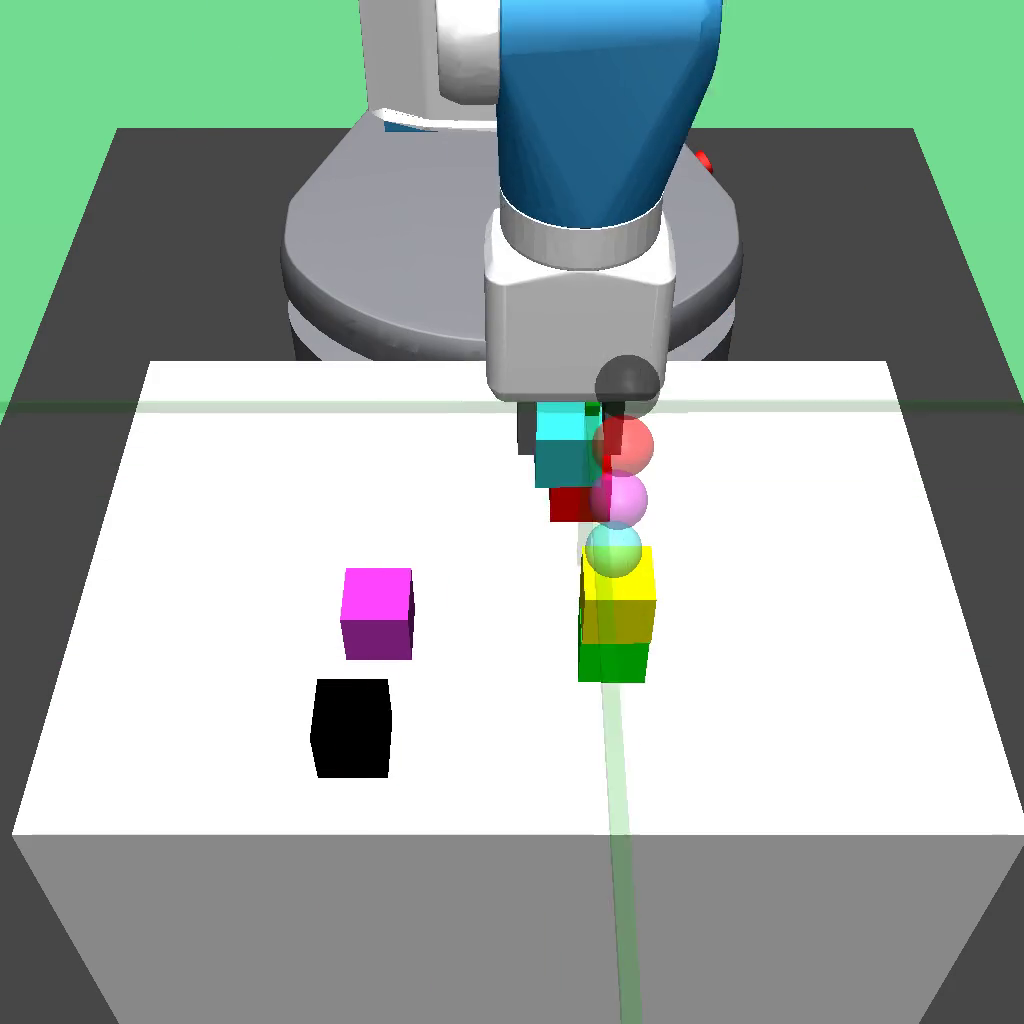}
\end{subfigure}
\begin{subfigure}[b]{0.124\linewidth}
    \centering
    \includegraphics[width=\linewidth]{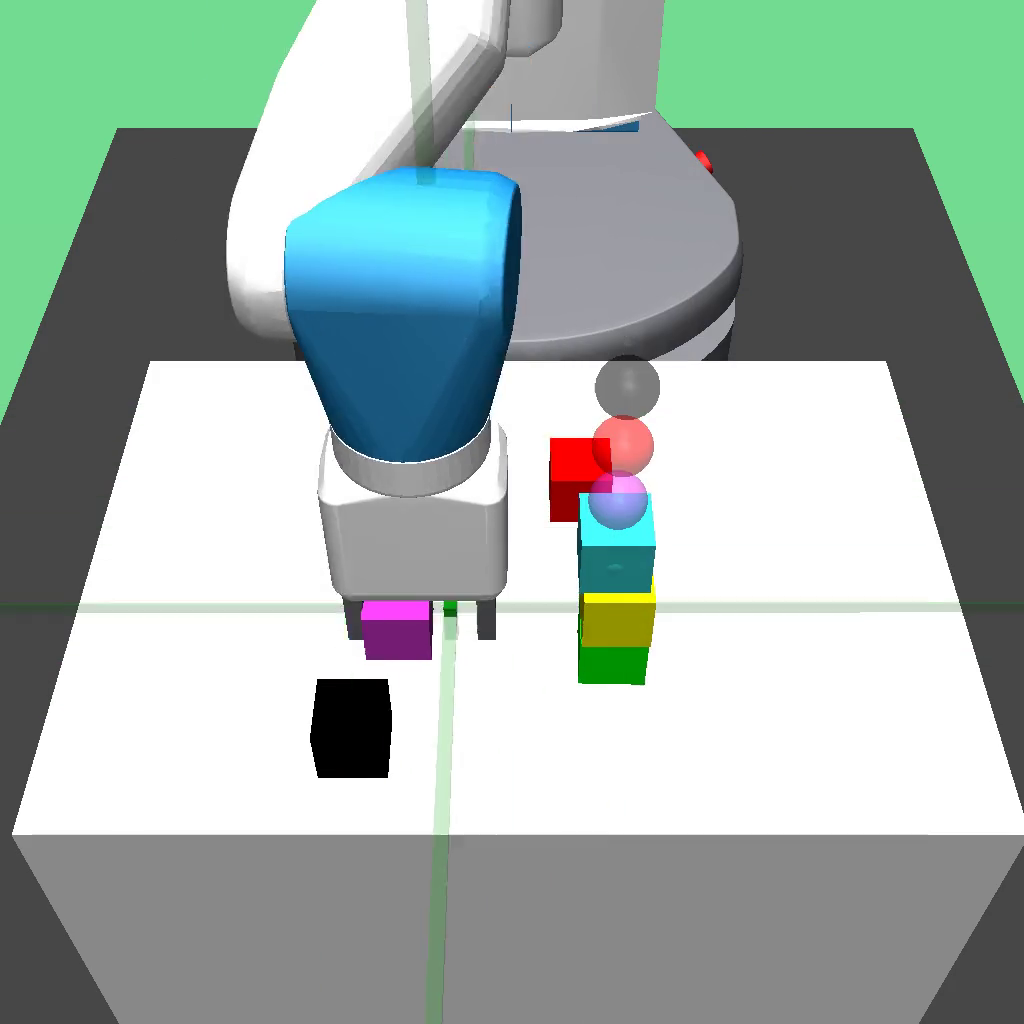}
\end{subfigure}
\begin{subfigure}[b]{0.124\linewidth}
    \centering
    \includegraphics[width=\linewidth]{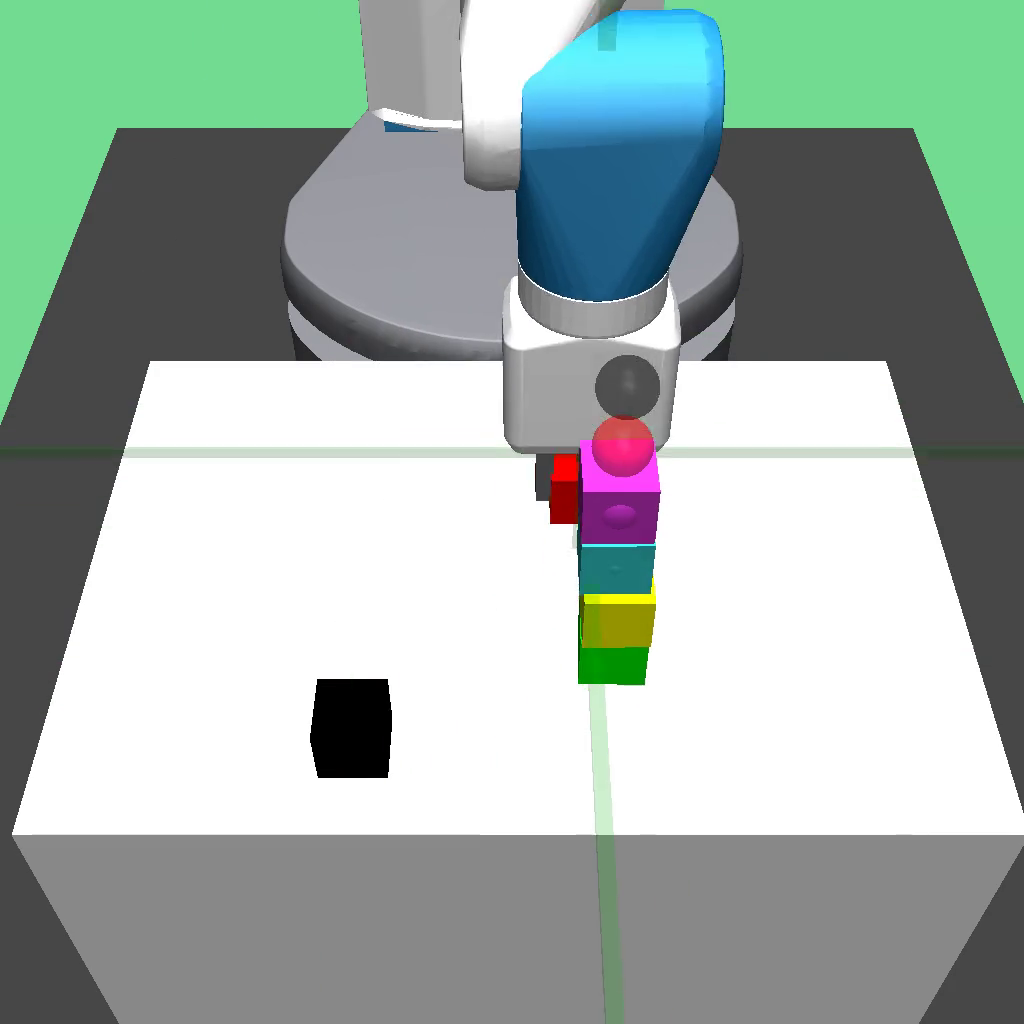}
\end{subfigure}
\begin{subfigure}[b]{0.124\linewidth}
    \includegraphics[width=\linewidth]{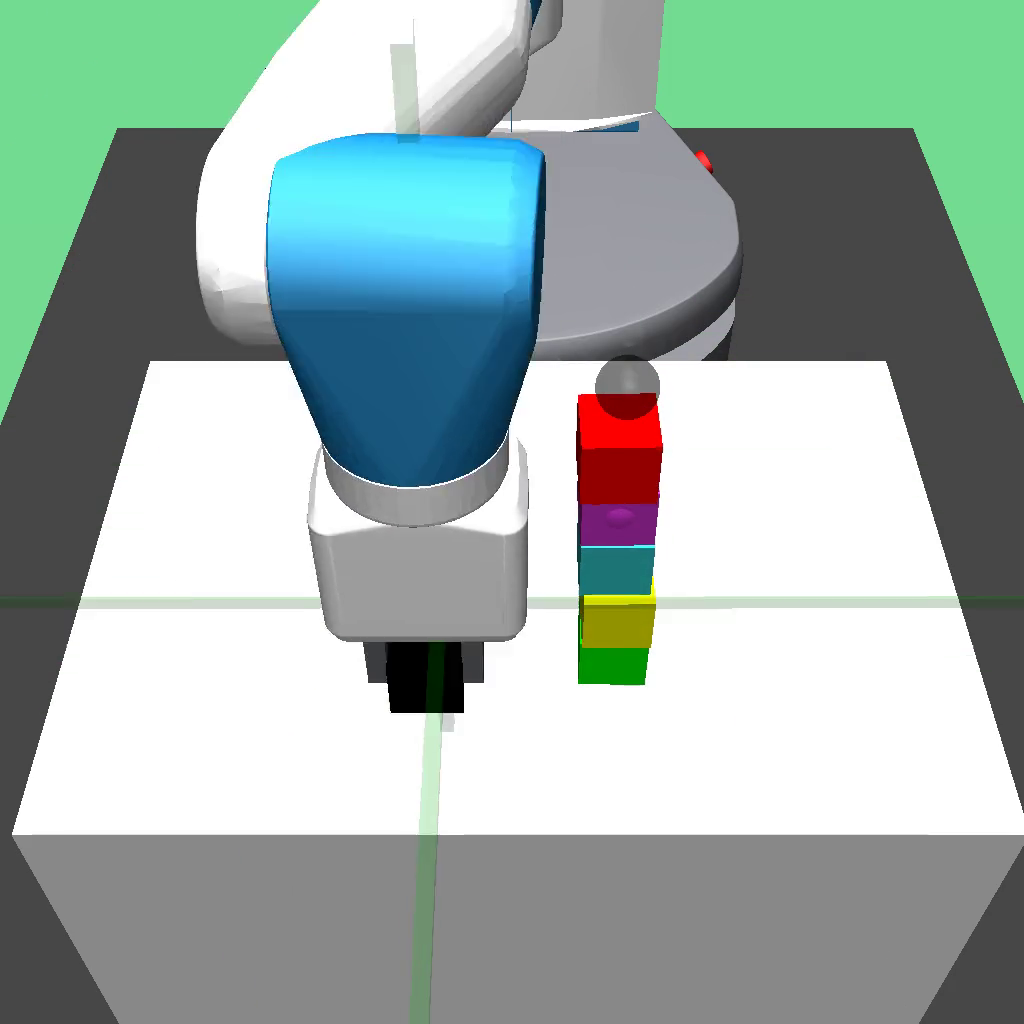}
\end{subfigure}
\begin{subfigure}[b]{0.124\linewidth}
    \includegraphics[width=\linewidth]{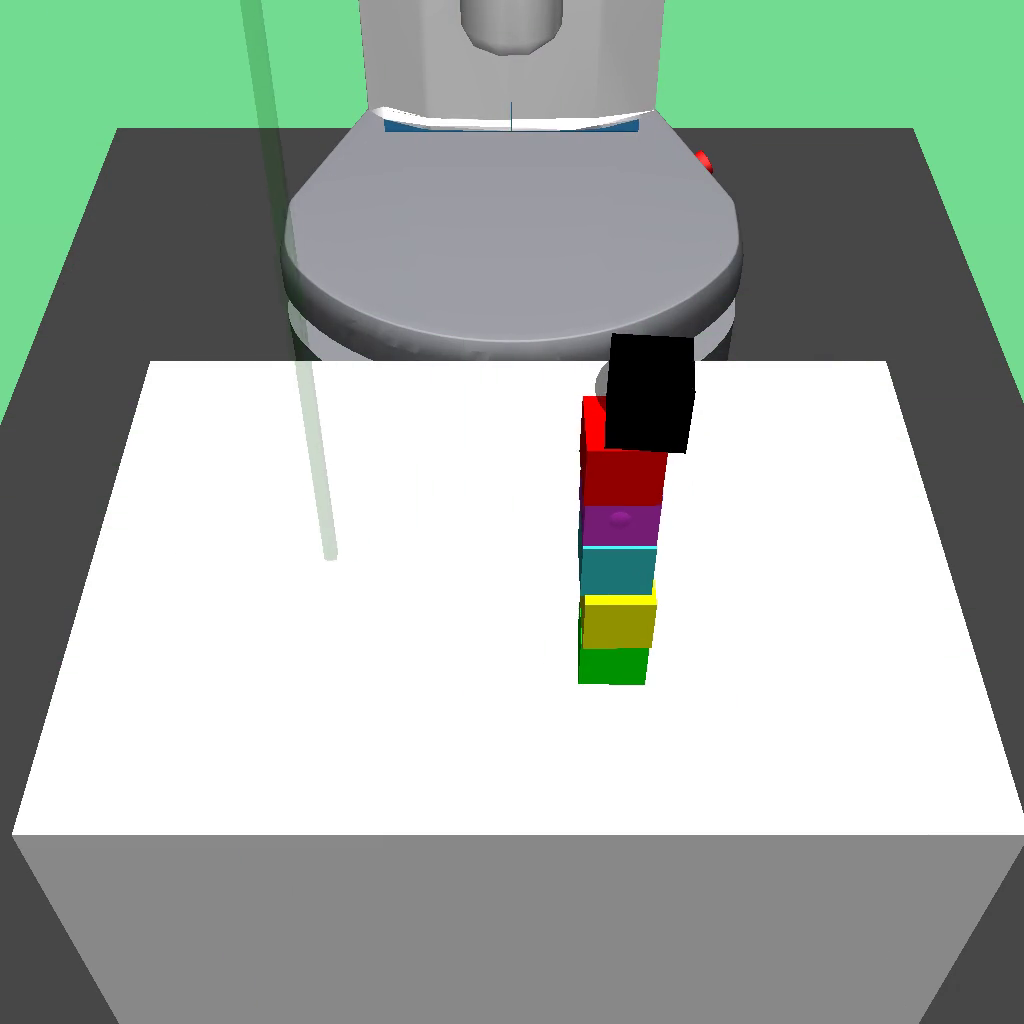}
\end{subfigure}

\begin{tikzpicture}
  \node (img)  {\includegraphics[scale=.184]{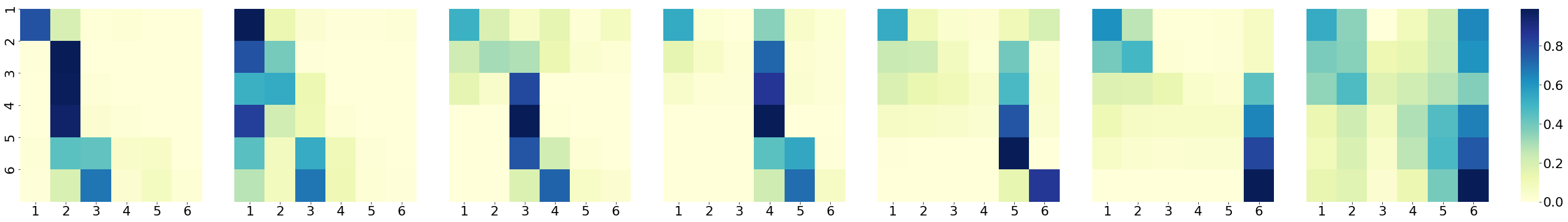}};
 \end{tikzpicture}
      \caption{Analysis of internal representation used by our system to solve the task of stacking six blocks into a single tower. The first row shows the key step of this task. The second row, shows for each of these key frames, a $6\times6$ matrix, whose $(i,j)^{th}$ entry represents the influence of feature representation of $j^{th}$ block on the $i^{th}$ block after three message passing rounds. The labelled index along the y-axis and x-axis correspond to the object ID. Object IDs 1 - 6 correspond to green, yellow, blue, pink, red, and black blocks in order. The attention map reveals that our agent pays attention to the sub-set of blocks relevant to solving the sub-problem (stacking one block) at hand. We speculate that such an attention map suggests that the agent has internally learnt to decompose the complex block stacking task into simpler sub-problems.}
      \label{fig:attention_heatmap}
      \vspace{-4mm}
\end{centering}
\end{figure*}

It is desirable to learn policies that are not only adept at the task they were trained on, but can be re-purposed for new and related tasks. If our \renn{} architecture indeed provides a good inductive bias, then it should be possible solve different block configuration tasks with high-accuracy. To test this, we evaluated the performance of the learned policy, without any fine-tuning on previously unseen block configurations (i.e. \textit{zero-shot generalization}) described in Section~\ref{sec:testing_details}. The results of this analysis are summarized in Figure~\ref{fig:zero_shot}. 

\noindent \minisection{Single Tower Evaluation:} Figure~\ref{fig:zero_shot} shows that a policy learned to stack $N$ blocks generalizes to stacking $N+1$ blocks without any training. The performance on stacking $N+k$ blocks, where $k>1$ drops significantly. One possible explanation is that it becomes progressively harder to stabilize larger number of blocks in a tower and the robot needs to substantially refine its strategy to stack more blocks. An analysis of failure modes is presented in Appendix B. 

\noindent \minisection{Multiple Towers Evaluation:} The previous experiments tested generalization to a larger number of blocks, but on the same task. To test if the learned policy generalizes to new tasks, we evaluated the performance on stacking multiple towers instead of a single tower. Results in Figure~\ref{fig:zero_shot}(b) show that the agent trained for stacking a single tower of $N$ blocks can successfully stack multiple towers $N+k$ blocks. The performance again drops for $k>2$. However, generalization to $k \geq 2$ is better on the multiple towers task as compared to the single tower task.  This suggest that while \renn~can generalize to a larger number of blocks than seen during training time, stacking a taller single tower without additional training is hard due to the difficulty of stabilizing a taller stack of blocks. 

\noindent \minisection{Pyramid Evaluation:}  To stress test our system further, we evaluated its performance on placing blocks in a pyramid configuration (see Figure \ref{fig:goal_configs}). Note that the robot never saw pyramids during training. Stacking blocks in pyramid is different than a tower, because now blocks may need to be balanced on two supporting blocks instead of only being stacked vertically.  Figure~\ref{fig:zero_shot}(c) shows that our system is able to generalize and manipulate larger number of blocks than seen in training into pyramid configurations. Interestingly, the agent trained on Single Tower 4 performs better on the difficult Pyramid 5 and Pyramid 6 tasks than the agent trained on Single Tower 6. One possible explanation is that the agent trained on taller tower overfits to stacking blocks vertically, and is less able to stack blocks at an horizontal offset, which is useful for the pyramid task. 

\noindent \minisection {Emergent Strategies:} The accompanying videos (\url{https://richardrl.github.io/relational-rl}) show that our agent automatically learns to push other blocks to grasp a particular block, grasps two blocks at a time and places them one by one to save time and other complex behaviors. These strategies emerge automatically as a consequence of optimizing a sparse reward function.  

To the best of our knowledge, ours is the first work that reports such zero-shot generalization on the block stacking task using RL. At the same time, we acknowledge, there is substantial room for improving the zero-shot results and the stacking performance. Some future directions are described in Section~\ref{sec:discussion}.  

\subsection{Analyzing the learned representations}

In order to gain insights into why \renn{} leads to faster convergence and better generalization, we visualized the attention patterns as the robot stacked six blocks (see Figure~\ref{fig:attention_heatmap}). The first row shows the key steps in tower stacking. Each column in second row is a $6\times6$ matrix ($E$). Each entry in the matrix, $e_{ij}$ represents the normalized relevance score of the $i^{th}$ block on the features of the $j^{th}$ block (see $w_{ij}$ defined in Section ~\ref{subsec:attention_graph_bg}) computed by the final layer. It can be seen that maximum attention is to paid to the block that is to be placed and the attention on existing blocks in the stack decreases from the top-most to bottom-most block. Such attention pattern suggests that our system has learned to focus on the blocks most relevant to current block being placed. This is interesting because it suggests that \renn{} has learned to decompose a complex problem into simpler sub-problems. We hypothesize that such decomposition is the reason why our system can learn from a task curricula and exhibits zero-shot generalization.    

\section{Discussion}
\label{sec:discussion}

We have presented a framework for learning long-horizon, sparse reward tasks using deep reinforcement learning, relational graph architecture and curriculum learning. While we are orders of magnitude more sample efficient than the the existing state-of-the-art, our method would still require a few dozen robots (corresponding to our 35 workers) and several days (assuming each action takes .25 seconds) of real world training to achieve a comparable environment step complexity. And while block stacking is representative of long-horizon, multi-object manipulation tasks, it is important to scale our method to tasks involving more complicated object geometries and more granular manipulation.

In the current work, the curriculum is manually designed and based on the principle that smaller sets of objects are easier to learn to manipulate than larger sets of objects. However, more complicated and effective curricula could exist along axes of variation beyond just the object cardinality, and discovering these curricula automatically is an interesting direction for future research. One point of concern with relational architectures is that the computation time is quadratic in the number of entities. Developing computationally efficient methods is therefore important to scale these methods to environments with much larger numbers of objects. Finally, while we have presented results from state observation, in the future we would like to scale our system to work from visual and other sensory observations. 

\section{Acknowledgements}
We acknowledge support from US Department of Defense, DARPA's Machine Common Sense Grant and the BAIR and BDD industrial consortia. We thank Amazon Web Services (AWS) for their generous support in the form of cloud credits.  We'd like to thank Vitchyr Pong, Kristian Hartikainen, Ashvin Nair and other members of the BAIR lab and the Improbable AI lab for helpful discussions during this project. 


\bibliographystyle{IEEEtran}
\bibliography{icra2019bib}{}

\begin{thebibliography}{10}
\providecommand{\url}[1]{#1}
\csname url@rmstyle\endcsname
\providecommand{\newblock}{\relax}
\providecommand{\bibinfo}[2]{#2}
\providecommand\BIBentrySTDinterwordspacing{\spaceskip=0pt\relax}
\providecommand\BIBentryALTinterwordstretchfactor{4}
\providecommand\BIBentryALTinterwordspacing{\spaceskip=\fontdimen2\font plus
\BIBentryALTinterwordstretchfactor\fontdimen3\font minus
  \fontdimen4\font\relax}
\providecommand\BIBforeignlanguage[2]{{%
\expandafter\ifx\csname l@#1\endcsname\relax
\typeout{** WARNING: IEEEtran.bst: No hyphenation pattern has been}%
\typeout{** loaded for the language `#1'. Using the pattern for}%
\typeout{** the default language instead.}%
\else
\language=\csname l@#1\endcsname
\fi
#2}}

\bibitem{hadfield2017inverse}
D.~Hadfield-Menell, S.~Milli, P.~Abbeel, S.~J. Russell, and A.~Dragan,
  ``Inverse reward design,'' in \emph{Advances in neural information processing
  systems}, 2017, pp. 6765--6774.

\bibitem{deisenroth2011learning}
M.~P. Deisenroth, C.~E. Rasmussen, and D.~Fox, ``Learning to control a low-cost
  manipulator using data-efficient reinforcement learning,'' \emph{Robotics
  Science and Systems}, pp. 57--64, 2011.

\bibitem{rusu2016progressive}
A.~A. Rusu, N.~C. Rabinowitz, G.~Desjardins, H.~Soyer, J.~Kirkpatrick,
  K.~Kavukcuoglu, R.~Pascanu, and R.~Hadsell, ``Progressive neural networks,''
  \emph{arXiv preprint arXiv:1606.04671}, 2016.

\bibitem{nair2017overcoming}
\BIBentryALTinterwordspacing
A.~Nair, B.~McGrew, M.~Andrychowicz, W.~Zaremba, and P.~Abbeel, ``Overcoming
  exploration in reinforcement learning with demonstrations,'' \emph{CoRR},
  vol. abs/1709.10089, 2017. [Online]. Available:
  \url{http://arxiv.org/abs/1709.10089}
\BIBentrySTDinterwordspacing

\bibitem{gu2017deep}
S.~Gu, E.~Holly, T.~Lillicrap, and S.~Levine, ``Deep reinforcement learning for
  robotic manipulation with asynchronous off-policy updates,'' in \emph{2017
  IEEE international conference on robotics and automation (ICRA)}.\hskip 1em
  plus 0.5em minus 0.4em\relax IEEE, 2017, pp. 3389--3396.

\bibitem{kormushev2010robot}
P.~Kormushev, S.~Calinon, and D.~G. Caldwell, ``Robot motor skill coordination
  with em-based reinforcement learning,'' in \emph{2010 IEEE/RSJ international
  conference on intelligent robots and systems}.\hskip 1em plus 0.5em minus
  0.4em\relax IEEE, 2010, pp. 3232--3237.

\bibitem{andrychowicz2018learning}
M.~Andrychowicz, B.~Baker, M.~Chociej, R.~Jozefowicz, B.~McGrew, J.~Pachocki,
  A.~Petron, M.~Plappert, G.~Powell, A.~Ray, \emph{et~al.}, ``Learning
  dexterous in-hand manipulation,'' \emph{arXiv preprint arXiv:1808.00177},
  2018.

\bibitem{rajeswaran2017learning}
A.~Rajeswaran, V.~Kumar, A.~Gupta, G.~Vezzani, J.~Schulman, E.~Todorov, and
  S.~Levine, ``Learning complex dexterous manipulation with deep reinforcement
  learning and demonstrations,'' \emph{arXiv preprint arXiv:1709.10087}, 2017.

\bibitem{hessel2018rainbow}
M.~Hessel, J.~Modayil, H.~Van~Hasselt, T.~Schaul, G.~Ostrovski, W.~Dabney,
  D.~Horgan, B.~Piot, M.~Azar, and D.~Silver, ``Rainbow: Combining improvements
  in deep reinforcement learning,'' in \emph{Thirty-Second AAAI Conference on
  Artificial Intelligence}, 2018.

\bibitem{sac}
\BIBentryALTinterwordspacing
T.~Haarnoja, A.~Zhou, P.~Abbeel, and S.~Levine, ``Soft actor-critic: Off-policy
  maximum entropy deep reinforcement learning with a stochastic actor,''
  \emph{CoRR}, vol. abs/1801.01290, 2018. [Online]. Available:
  \url{http://arxiv.org/abs/1801.01290}
\BIBentrySTDinterwordspacing

\bibitem{schulman2017proximal}
J.~Schulman, F.~Wolski, P.~Dhariwal, A.~Radford, and O.~Klimov, ``Proximal
  policy optimization algorithms,'' \emph{arXiv preprint arXiv:1707.06347},
  2017.

\bibitem{levineFDA15}
S.~Levine, C.~Finn, T.~Darrell, and P.~Abbeel, ``End-to-end training of deep
  visuomotor policies,'' \emph{JMLR}, 2016.

\bibitem{suttonoptions1999}
R.~Sutton, D.~Precup, and S.~Singh, ``Between {MDP}s and semi-{MDP}s: {A}
  framework for temporal abstraction in reinforcement learning,''
  \emph{Artificial Intelligence}, vol. 112, pp. 181--211, 1999.

\bibitem{schmidhuber2010formal}
J.~Schmidhuber, ``Formal theory of creativity, fun, and intrinsic motivation
  (1990--2010),'' \emph{IEEE Transactions on Autonomous Mental Development},
  2010.

\bibitem{lopes2012exploration}
M.~Lopes, T.~Lang, M.~Toussaint, and P.-Y. Oudeyer, ``Exploration in
  model-based reinforcement learning by empirically estimating learning
  progress,'' in \emph{NIPS}, 2012.

\bibitem{oudeyer2009intrinsic}
P.-Y. Oudeyer and F.~Kaplan, ``What is intrinsic motivation? a typology of
  computational approaches,'' \emph{Frontiers in neurorobotics}, 2009.

\bibitem{pathak2017curiosity}
D.~Pathak, P.~Agrawal, A.~A. Efros, and T.~Darrell, ``Curiosity-driven
  exploration by self-supervised prediction,'' in \emph{Proceedings of the IEEE
  Conference on Computer Vision and Pattern Recognition Workshops}, 2017, pp.
  16--17.

\bibitem{sukhbaatar2018intrinsic}
\BIBentryALTinterwordspacing
S.~Sukhbaatar, Z.~Lin, I.~Kostrikov, G.~Synnaeve, A.~Szlam, and R.~Fergus,
  ``Intrinsic motivation and automatic curricula via asymmetric self-play,'' in
  \emph{International Conference on Learning Representations}, 2018. [Online].
  Available: \url{https://openreview.net/forum?id=SkT5Yg-RZ}
\BIBentrySTDinterwordspacing

\bibitem{zambaldi2019deep}
V.~Zambaldi, D.~Raposo, A.~Santoro, V.~Bapst, Y.~Li, I.~Babuschkin, K.~Tuyls,
  D.~Reichert, T.~Lillicrap, E.~Lockhart, \emph{et~al.}, ``Deep reinforcement
  learning with relational inductive biases,'' \emph{International Conference
  on Learning Representations}, 2019.

\bibitem{wang2018nervenet}
\BIBentryALTinterwordspacing
T.~Wang, R.~Liao, J.~Ba, and S.~Fidler, ``Nervenet: Learning structured policy
  with graph neural networks,'' in \emph{International Conference on Learning
  Representations}, 2018. [Online]. Available:
  \url{https://openreview.net/forum?id=S1sqHMZCb}
\BIBentrySTDinterwordspacing

\bibitem{pathak19assemblies}
D.~Pathak, C.~Lu, T.~Darrell, P.~Isola, and A.~A. Efros, ``Learning to control
  self- assembling morphologies: A study of generalization via modularity,'' in
  \emph{arXiv preprint arXiv:1902.05546}, 2019.

\bibitem{donahue2014decaf}
J.~Donahue, Y.~Jia, O.~Vinyals, J.~Hoffman, N.~Zhang, E.~Tzeng, and T.~Darrell,
  ``Decaf: A deep convolutional activation feature for generic visual
  recognition,'' in \emph{International conference on machine learning}, 2014,
  pp. 647--655.

\bibitem{agrawal2016learning}
P.~Agrawal, A.~Nair, P.~Abbeel, J.~Malik, and S.~Levine, ``Learning to poke by
  poking: Experiential learning of intuitive physics,'' \emph{NIPS}, 2016.

\bibitem{ren2015faster}
S.~Ren, K.~He, R.~Girshick, and J.~Sun, ``Faster r-cnn: Towards real-time
  object detection with region proposal networks,'' in \emph{Advances in neural
  information processing systems}, 2015, pp. 91--99.

\bibitem{zhang2018study}
C.~Zhang, O.~Vinyals, R.~Munos, and S.~Bengio, ``A study on overfitting in deep
  reinforcement learning,'' \emph{arXiv preprint arXiv:1804.06893}, 2018.

\bibitem{cobbe2018quantifying}
K.~Cobbe, O.~Klimov, C.~Hesse, T.~Kim, and J.~Schulman, ``Quantifying
  generalization in reinforcement learning,'' \emph{arXiv preprint
  arXiv:1812.02341}, 2018.

\bibitem{justesen2018illuminating}
N.~Justesen, R.~R. Torrado, P.~Bontrager, A.~Khalifa, J.~Togelius, and S.~Risi,
  ``Illuminating generalization in deep reinforcement learning through
  procedural level generation,'' \emph{arXiv preprint arXiv:1806.10729}, 2018.

\bibitem{nichol2018first}
A.~Nichol, J.~Achiam, and J.~Schulman, ``On first-order meta-learning
  algorithms,'' \emph{arXiv preprint arXiv:1803.02999}, 2018.

\bibitem{ghosh2017divide}
D.~Ghosh, A.~Singh, A.~Rajeswaran, V.~Kumar, and S.~Levine,
  ``Divide-and-conquer reinforcement learning,'' \emph{arXiv preprint
  arXiv:1711.09874}, 2017.

\bibitem{bengio2009curriculum}
Y.~Bengio, J.~Louradour, R.~Collobert, and J.~Weston, ``Curriculum learning,''
  in \emph{Proceedings of the 26th annual international conference on machine
  learning}.\hskip 1em plus 0.5em minus 0.4em\relax ACM, 2009, pp. 41--48.

\bibitem{tobin2017domain}
J.~Tobin, R.~Fong, A.~Ray, J.~Schneider, W.~Zaremba, and P.~Abbeel, ``Domain
  randomization for transferring deep neural networks from simulation to the
  real world,'' in \emph{2017 IEEE/RSJ International Conference on Intelligent
  Robots and Systems (IROS)}.\hskip 1em plus 0.5em minus 0.4em\relax IEEE,
  2017, pp. 23--30.

\bibitem{kaelbling2001learning}
L.~P. Kaelbling, T.~Oates, N.~Hernandez, and S.~Finney, ``Learning in worlds
  with objects,'' in \emph{Working Notes of the AAAI Stanford Spring Symposium
  on Learning Grounded Representations}, 2001, pp. 31--36.

\bibitem{dzeroski2001}
S.~Džeroski, L.~De~Raedt, and K.~Driessens, ``Relational reinforcement
  learning,'' \emph{Machine Learning}, vol.~43, no.~1, p. 7–52, Apr 2001.

\bibitem{van2002relational}
M.~Van~Otterlo, ``Relational representations in reinforcement learning: Review
  and open problems,'' in \emph{Proceedings of the ICML}, vol.~2, 2002.

\bibitem{Russell_Norvig_2009}
S.~Russell and P.~Norvig, \emph{Artificial Intelligence: A Modern Approach},
  3rd~ed.\hskip 1em plus 0.5em minus 0.4em\relax Prentice Hall Press, 2009.

\bibitem{bruna14}
J.~Bruna, W.~Zaremba, A.~Szlam, and Y.~Lecun, ``\BIBforeignlanguage{English
  (US)}{Spectral networks and locally connected networks on graphs},'' in
  \emph{\BIBforeignlanguage{English (US)}{International Conference on Learning
  Representations (ICLR2014), CBLS, April 2014}}, 2014.

\bibitem{defferrard16}
\BIBentryALTinterwordspacing
M.~Defferrard, X.~Bresson, and P.~Vandergheynst, ``Convolutional neural
  networks on graphs with fast localized spectral filtering,'' in
  \emph{Advances in Neural Information Processing Systems 29}, D.~D. Lee,
  M.~Sugiyama, U.~V. Luxburg, I.~Guyon, and R.~Garnett, Eds.\hskip 1em plus
  0.5em minus 0.4em\relax Curran Associates, Inc., 2016, pp. 3844--3852.
  [Online]. Available:
  \url{http://papers.nips.cc/paper/6081-convolutional-neural-networks-on-graphs-with-fast-localized-spectral-filtering.pdf}
\BIBentrySTDinterwordspacing

\bibitem{kipf2016semi}
T.~N. Kipf and M.~Welling, ``Semi-supervised classification with graph
  convolutional networks,'' \emph{arXiv preprint arXiv:1609.02907}, 2016.

\bibitem{structuralrnn16}
\BIBentryALTinterwordspacing
A.~Jain, A.~R. Zamir, S.~Savarese, and A.~Saxena, ``Structural-rnn: Deep
  learning on spatio-temporal graphs,'' \emph{CoRR}, vol. abs/1511.05298, 2015.
  [Online]. Available: \url{http://arxiv.org/abs/1511.05298}
\BIBentrySTDinterwordspacing

\bibitem{transformer}
\BIBentryALTinterwordspacing
A.~Vaswani, N.~Shazeer, N.~Parmar, J.~Uszkoreit, L.~Jones, A.~N. Gomez,
  L.~Kaiser, and I.~Polosukhin, ``Attention is all you need,'' \emph{CoRR},
  vol. abs/1706.03762, 2017. [Online]. Available:
  \url{http://arxiv.org/abs/1706.03762}
\BIBentrySTDinterwordspacing

\bibitem{battaglia16}
\BIBentryALTinterwordspacing
P.~Battaglia, R.~Pascanu, M.~Lai, D.~J. Rezende, and K.~kavukcuoglu,
  ``Interaction networks for learning about objects, relations and physics,''
  in \emph{Proceedings of the 30th International Conference on Neural
  Information Processing Systems}, ser. NIPS'16.\hskip 1em plus 0.5em minus
  0.4em\relax USA: Curran Associates Inc., 2016, pp. 4509--4517. [Online].
  Available: \url{http://dl.acm.org/citation.cfm?id=3157382.3157601}
\BIBentrySTDinterwordspacing

\bibitem{oneshotimitationlearning}
\BIBentryALTinterwordspacing
Y.~Duan, M.~Andrychowicz, B.~Stadie, O.~Jonathan~Ho, J.~Schneider,
  I.~Sutskever, P.~Abbeel, and W.~Zaremba, ``One-shot imitation learning,'' in
  \emph{Advances in Neural Information Processing Systems 30}, I.~Guyon, U.~V.
  Luxburg, S.~Bengio, H.~Wallach, R.~Fergus, S.~Vishwanathan, and R.~Garnett,
  Eds.\hskip 1em plus 0.5em minus 0.4em\relax Curran Associates, Inc., 2017,
  pp. 1087--1098. [Online]. Available:
  \url{http://papers.nips.cc/paper/6709-one-shot-imitation-learning.pdf}
\BIBentrySTDinterwordspacing

\bibitem{janner2019reasoning}
M.~Janner, S.~Levine, W.~T. Freeman, J.~B. Tenenbaum, C.~Finn, and J.~Wu,
  ``Reasoning about physical interactions with object-oriented prediction and
  planning,'' in \emph{International Conference on Learning Representations},
  2019.

\bibitem{elman1993}
J.~L. Elman, ``Learning and development in neural networks: the importance of
  starting small,'' \emph{Cognition}, vol.~48, no.~1, p. 71–99, 1993.

\bibitem{Bengio2009}
\BIBentryALTinterwordspacing
Y.~Bengio, J.~Louradour, R.~Collobert, and J.~Weston, ``Curriculum learning,''
  in \emph{Proceedings of the 26th Annual International Conference on Machine
  Learning}, ser. ICML '09.\hskip 1em plus 0.5em minus 0.4em\relax New York,
  NY, USA: ACM, 2009, pp. 41--48. [Online]. Available:
  \url{http://doi.acm.org/10.1145/1553374.1553380}
\BIBentrySTDinterwordspacing

\bibitem{zaremba14}
\BIBentryALTinterwordspacing
W.~Zaremba and I.~Sutskever, ``Learning to execute,'' \emph{CoRR}, vol.
  abs/1410.4615, 2014. [Online]. Available:
  \url{http://arxiv.org/abs/1410.4615}
\BIBentrySTDinterwordspacing

\bibitem{graves17}
\BIBentryALTinterwordspacing
A.~Graves, M.~G. Bellemare, J.~Menick, R.~Munos, and K.~Kavukcuoglu,
  ``Automated curriculum learning for neural networks,'' \emph{CoRR}, vol.
  abs/1704.03003, 2017. [Online]. Available:
  \url{http://arxiv.org/abs/1704.03003}
\BIBentrySTDinterwordspacing

\bibitem{goalgan17}
\BIBentryALTinterwordspacing
D.~Held, X.~Geng, C.~Florensa, and P.~Abbeel, ``Automatic goal generation for
  reinforcement learning agents,'' \emph{CoRR}, vol. abs/1705.06366, 2017.
  [Online]. Available: \url{http://arxiv.org/abs/1705.06366}
\BIBentrySTDinterwordspacing

\bibitem{florensa2017reverse}
C.~Florensa, D.~Held, M.~Wulfmeier, M.~Zhang, and P.~Abbeel, ``Reverse
  curriculum generation for reinforcement learning,'' \emph{arXiv preprint
  arXiv:1707.05300}, 2017.

\bibitem{duan2017one}
Y.~Duan, M.~Andrychowicz, B.~Stadie, O.~J. Ho, J.~Schneider, I.~Sutskever,
  P.~Abbeel, and W.~Zaremba, ``One-shot imitation learning,'' in \emph{Advances
  in neural information processing systems}, 2017, pp. 1087--1098.

\bibitem{popov2018dataefficient}
\BIBentryALTinterwordspacing
I.~Popov, N.~Heess, T.~P. Lillicrap, R.~Hafner, G.~Barth-Maron, M.~Vecerik,
  T.~Lampe, T.~Erez, Y.~Tassa, and M.~Riedmiller, ``Data-efficient deep
  reinforcement learning for dexterous manipulation,'' 2018. [Online].
  Available: \url{https://openreview.net/forum?id=SJdCUMZAW}
\BIBentrySTDinterwordspacing

\bibitem{kroemer2018kernel}
O.~Kroemer, S.~Leischnig, S.~Luettgen, and J.~Peters, ``A kernel-based approach
  to learning contact distributions for robot manipulation tasks,''
  \emph{Autonomous Robots}, vol.~42, no.~3, pp. 581--600, 2018.

\bibitem{toussaint2015logic}
M.~Toussaint, ``Logic-geometric programming: An optimization-based approach to
  combined task and motion planning,'' in \emph{Twenty-Fourth International
  Joint Conference on Artificial Intelligence}, 2015.

\bibitem{feudalnets}
\BIBentryALTinterwordspacing
A.~S. Vezhnevets, S.~Osindero, T.~Schaul, N.~Heess, M.~Jaderberg, D.~Silver,
  and K.~Kavukcuoglu, ``Feudal networks for hierarchical reinforcement
  learning,'' \emph{CoRR}, vol. abs/1703.01161, 2017. [Online]. Available:
  \url{http://arxiv.org/abs/1703.01161}
\BIBentrySTDinterwordspacing

\bibitem{dayan1993feudal}
P.~Dayan and G.~E. Hinton, ``Feudal reinforcement learning,'' in \emph{Advances
  in neural information processing systems}, 1993, pp. 271--278.

\bibitem{maxq}
T.~G. Dietterich, ``The maxq method for hierarchical reinforcement learning,''
  in \emph{In Proceedings of the Fifteenth International Conference on Machine
  Learning}.\hskip 1em plus 0.5em minus 0.4em\relax Morgan Kaufmann, 1998, pp.
  118--126.

\bibitem{bacon16}
\BIBentryALTinterwordspacing
P.~Bacon, J.~Harb, and D.~Precup, ``The option-critic architecture,''
  \emph{CoRR}, vol. abs/1609.05140, 2016. [Online]. Available:
  \url{http://arxiv.org/abs/1609.05140}
\BIBentrySTDinterwordspacing

\bibitem{modulatedpolicyhierarchies}
\BIBentryALTinterwordspacing
A.~Pashevich, D.~Hafner, J.~Davidson, R.~Sukthankar, and C.~Schmid, ``Modulated
  policy hierarchies,'' \emph{CoRR}, vol. abs/1812.00025, 2018. [Online].
  Available: \url{http://arxiv.org/abs/1812.00025}
\BIBentrySTDinterwordspacing

\bibitem{plappert2018multi}
M.~Plappert, M.~Andrychowicz, A.~Ray, B.~McGrew, B.~Baker, G.~Powell,
  J.~Schneider, J.~Tobin, M.~Chociej, P.~Welinder, \emph{et~al.}, ``Multi-goal
  reinforcement learning: Challenging robotics environments and request for
  research,'' \emph{arXiv preprint arXiv:1802.09464}, 2018.

\bibitem{mujoco}
\BIBentryALTinterwordspacing
E.~Todorov, T.~Erez, and Y.~Tassa, ``Mujoco: A physics engine for model-based
  control.'' in \emph{IROS}.\hskip 1em plus 0.5em minus 0.4em\relax IEEE, 2012,
  pp. 5026--5033. [Online]. Available:
  \url{http://dblp.uni-trier.de/db/conf/iros/iros2012.html#TodorovET12}
\BIBentrySTDinterwordspacing

\bibitem{watkins1992q}
C.~J. Watkins and P.~Dayan, ``Q-learning,'' \emph{Machine learning}, vol.~8,
  no. 3-4, pp. 279--292, 1992.

\bibitem{sutton1998}
\BIBentryALTinterwordspacing
R.~S. Sutton and A.~G. Barto, \emph{Reinforcement Learning: An Introduction},
  2nd~ed.\hskip 1em plus 0.5em minus 0.4em\relax The MIT Press, 2018. [Online].
  Available: \url{http://incompleteideas.net/book/the-book-2nd.html}
\BIBentrySTDinterwordspacing

\bibitem{schaul2015universal}
T.~Schaul, D.~Horgan, K.~Gregor, and D.~Silver, ``Universal value function
  approximators,'' in \emph{International Conference on Machine Learning},
  2015, pp. 1312--1320.

\bibitem{her}
\BIBentryALTinterwordspacing
M.~Andrychowicz, F.~Wolski, A.~Ray, J.~Schneider, R.~Fong, P.~Welinder,
  B.~McGrew, J.~Tobin, O.~Pieter~Abbeel, and W.~Zaremba, ``Hindsight experience
  replay,'' in \emph{Advances in Neural Information Processing Systems 30},
  I.~Guyon, U.~V. Luxburg, S.~Bengio, H.~Wallach, R.~Fergus, S.~Vishwanathan,
  and R.~Garnett, Eds.\hskip 1em plus 0.5em minus 0.4em\relax Curran
  Associates, Inc., 2017, pp. 5048--5058. [Online]. Available:
  \url{http://papers.nips.cc/paper/7090-hindsight-experience-replay.pdf}
\BIBentrySTDinterwordspacing

\bibitem{lillicrap2015continuous}
T.~P. Lillicrap, J.~J. Hunt, A.~Pritzel, N.~Heess, T.~Erez, Y.~Tassa,
  D.~Silver, and D.~Wierstra, ``Continuous control with deep reinforcement
  learning,'' \emph{ICLR}, 2016.

\end{thebibliography}

\newpage

\let\oldtwocolumn\twocolumn
\renewcommand\twocolumn[1][]{%
    \oldtwocolumn[{#1}{
    \begin{flushleft}
          \includegraphics[width=0.98\textwidth]{images/icra2019paperheader.png}
          \captionof{figure}{We present a simple yet effective reinforcement learning system that can stack 6 blocks in a tower without requiring any demonstrations or task-specific assumptions. Our method also exhibits zero-shot generalization and is capable of configuring blocks into previously unseen configurations of multiple towers and pyramids without any training (last two rows). See the videos here: \url{https://richardrl.github.io/relational-rl}}
          \label{fig:fig8}
            \end{flushleft}
    }]
}

\begin{appendix}
\renewcommand{\thefigure}{A.\arabic{figure}}
\renewcommand{\thetable}{A.\arabic{table}}
\setcounter{figure}{0}
\setcounter{table}{0}

In this section, we provide additional details regarding the experimental setup and results.

\subsection{Hyperparameter Tables}
\begin{table}[h!]
    \caption{Experiment Hyperparameters.}
    \label{h1}
    \begin{center}
    \begin{tabular}{|c||c|}
    \hline
    Parameter & Setting\\  \hline
    Number of workers & 35 \\ \hline
    Replay buffer max size & 1E5 \\ \hline
    Optimizer & Adam \\ \hline
    Learning rate & 3E-4 \\ \hline
    Epsilon for uniform action sampling & .1 \\ \hline
    Discount factor & .98 \\ \hline
    Batch size & 256 \\ \hline
    HER fraction of re-labelled goals & .8 \\ \hline
    SAC target entropy & 4 \\ \hline
    \end{tabular}
    \end{center}
\end{table}

\begin{table}[ht!]
    \caption{ReNN Architecture Hyperparameters.}
    \begin{center}
    \begin{tabular}{|c||c|}
    \hline
    Parameter & Setting\\  \hline
    Embedding dimension & 64 \\ \hline
    Number of graph modules & 3 \\ \hline
    Graph module weight sharing & False \\ \hline
    Post-readout MLP hidden layers & 3 \\ \hline
    Input normalization & Shared mean/stddev for each block \\ \hline
    Activation function & Leaky ReLU \\ \hline
    Activation function (attention) & tanh \\ \hline
    \end{tabular}
    \end{center}
\end{table}

\subsection{Successful Configurations for Goal Types}
\begin{figure}[h!]
\begin{subfigure}[t]{.93\linewidth}
\begin{subfigure}[b]{0.31\linewidth}
    \centering
    \includegraphics[width=1.0\linewidth]{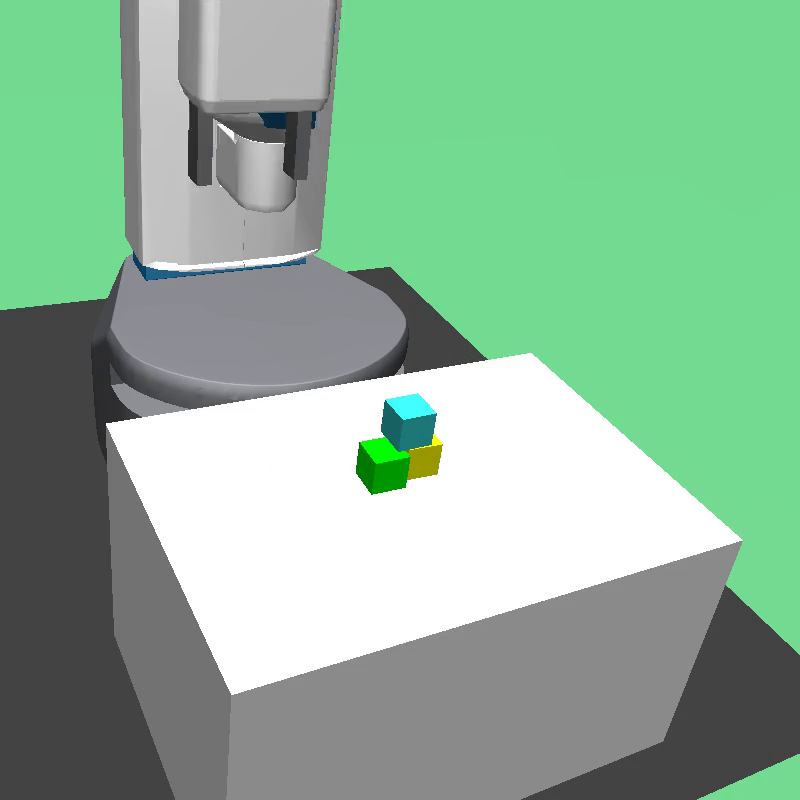}
\end{subfigure}
\begin{subfigure}[b]{0.31\linewidth}
    \centering
    \includegraphics[width=1.0\linewidth]{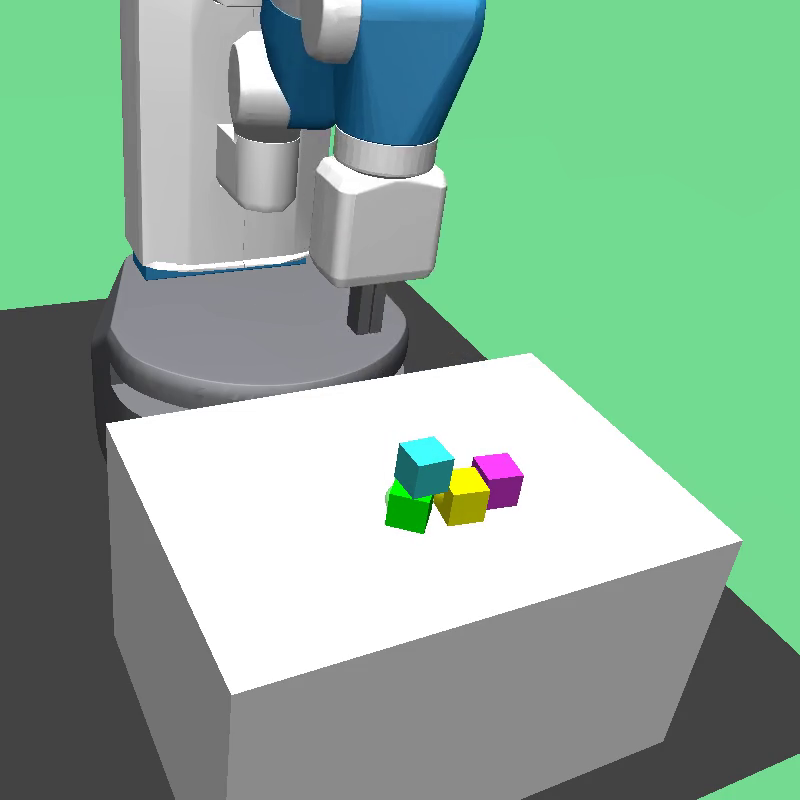}
\end{subfigure}
\begin{subfigure}[b]{0.31\linewidth}
    \centering
    \includegraphics[width=1.0\linewidth]{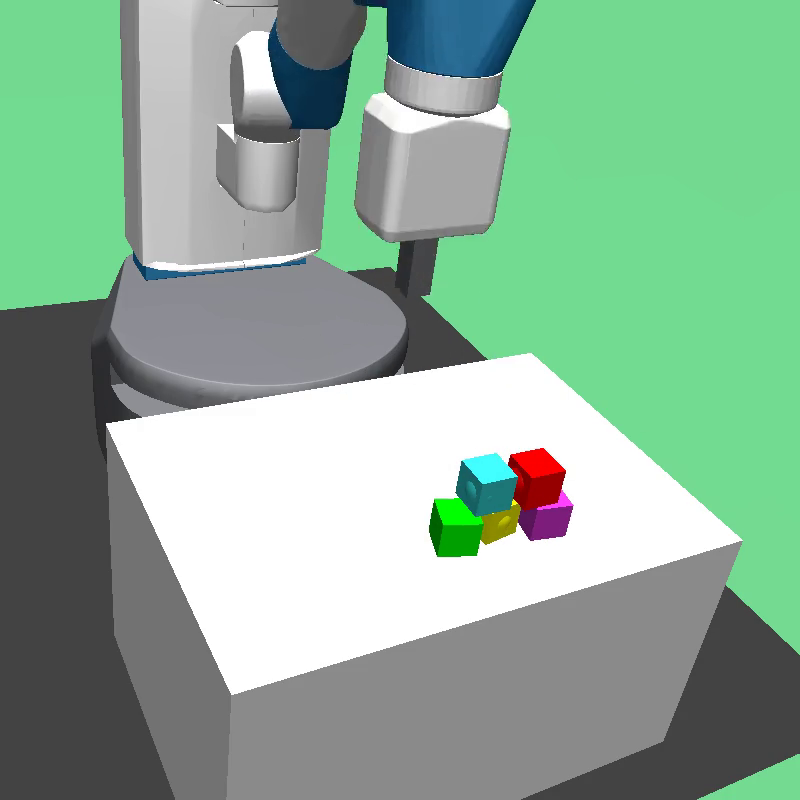}
\end{subfigure}
\\[1ex]
\begin{subfigure}[b]{0.31\linewidth}
    \centering
    \includegraphics[width=1.0\linewidth]{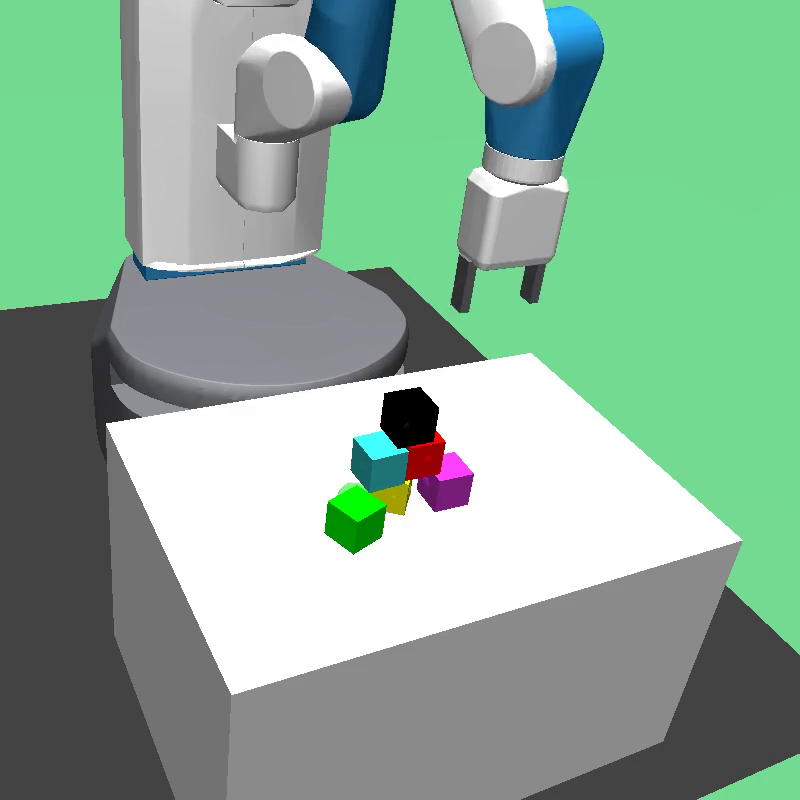}
\end{subfigure}
\begin{subfigure}[b]{0.31\linewidth}
    \centering
    \includegraphics[width=1.0\linewidth]{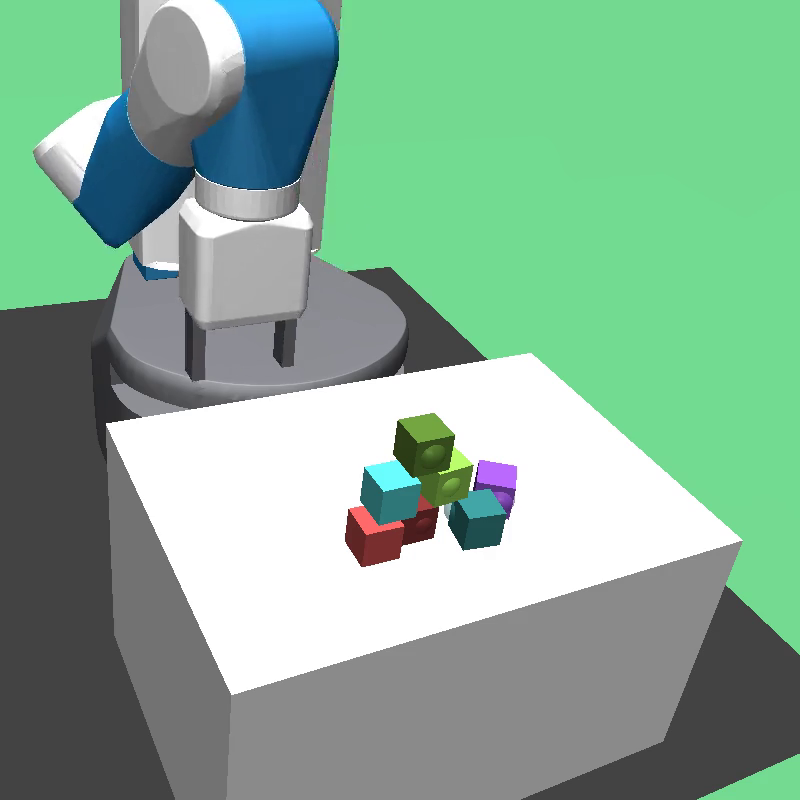}
\end{subfigure}
\caption{}
\end{subfigure}

\vskip 3pt

\begin{subfigure}[t]{.93\linewidth}
\begin{subfigure}[b]{0.31\linewidth}
    \centering
    \includegraphics[width=1.0\linewidth]{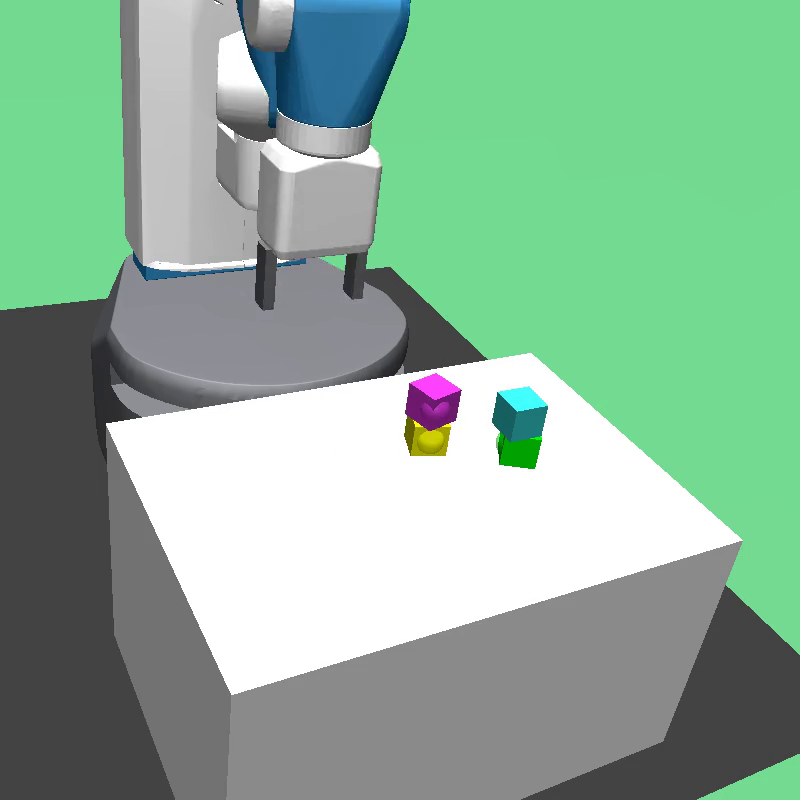}
\end{subfigure}
\begin{subfigure}[b]{0.31\linewidth}
    \centering
    \includegraphics[width=1.0\linewidth]{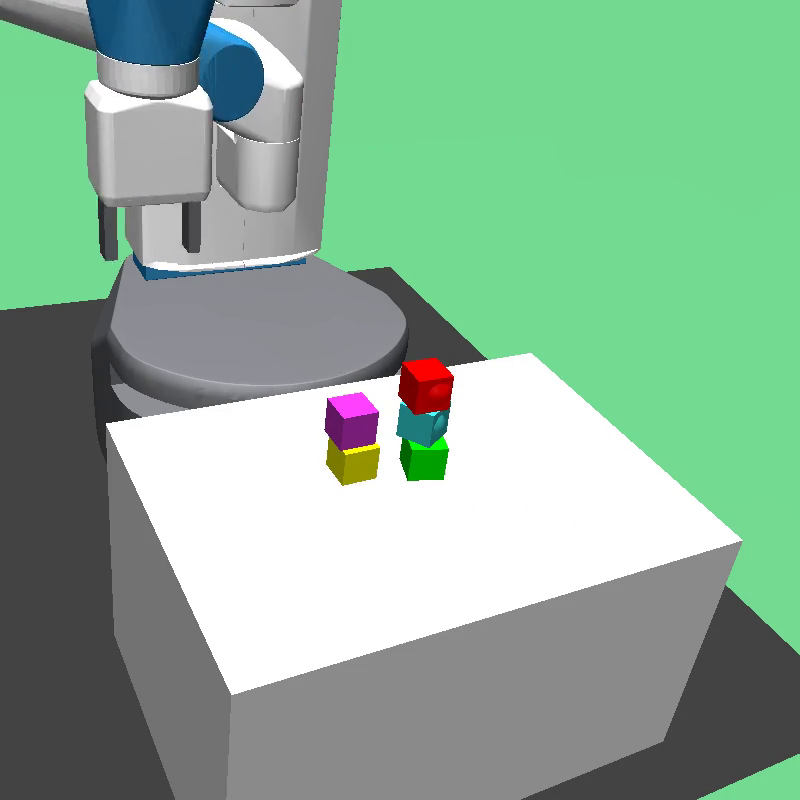}
\end{subfigure}
\begin{subfigure}[b]{0.31\linewidth}
    \centering
    \includegraphics[width=1.0\linewidth]{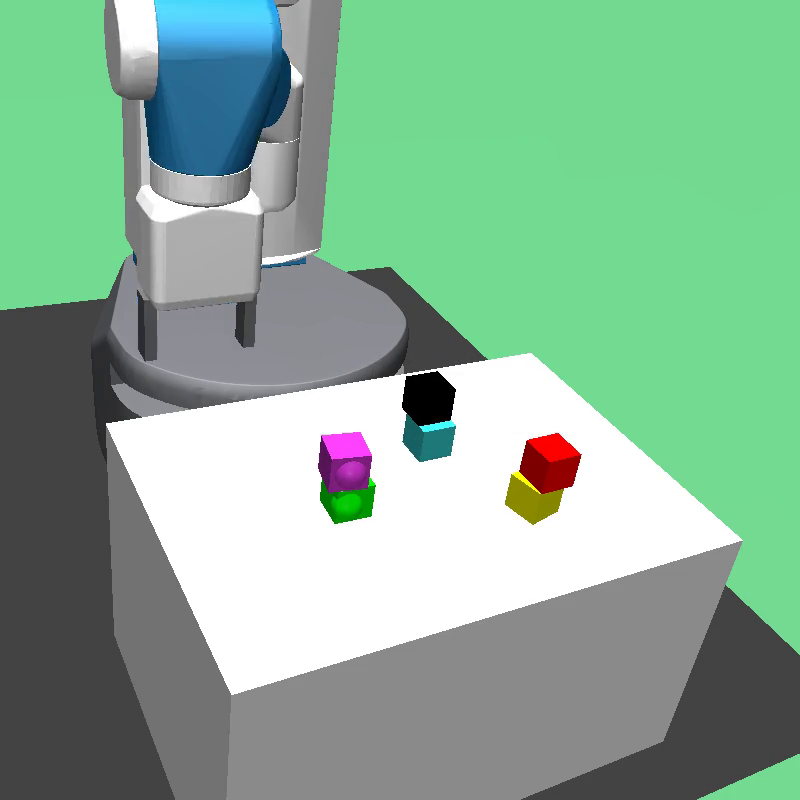}
\end{subfigure}
\\[1ex]
\begin{subfigure}[b]{0.31\linewidth}
    \centering
    \includegraphics[width=1.0\linewidth]{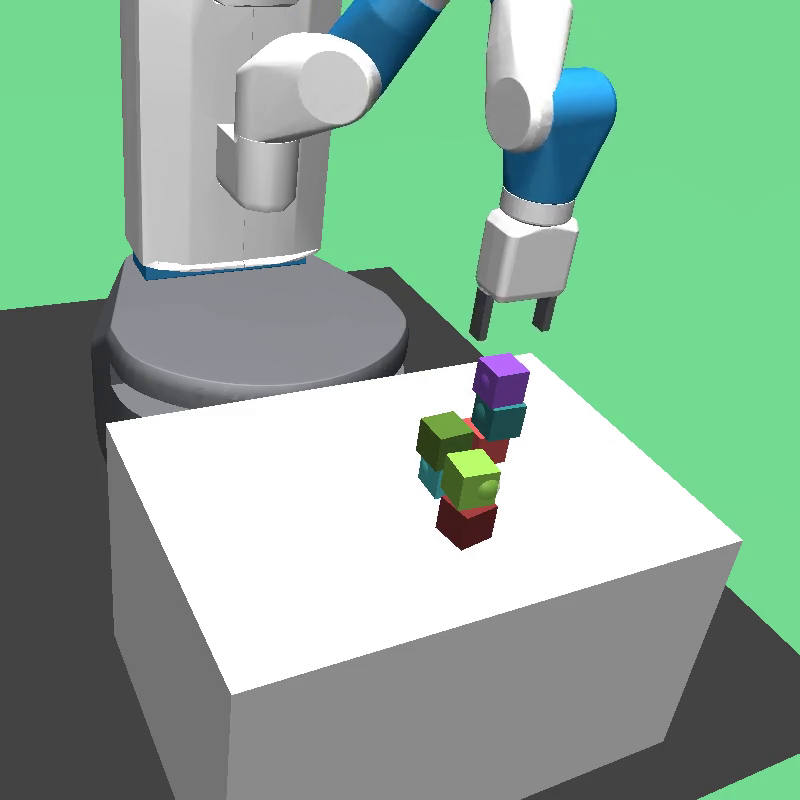}
\end{subfigure}
\begin{subfigure}[b]{0.31\linewidth}
    \centering
    \includegraphics[width=1.0\linewidth]{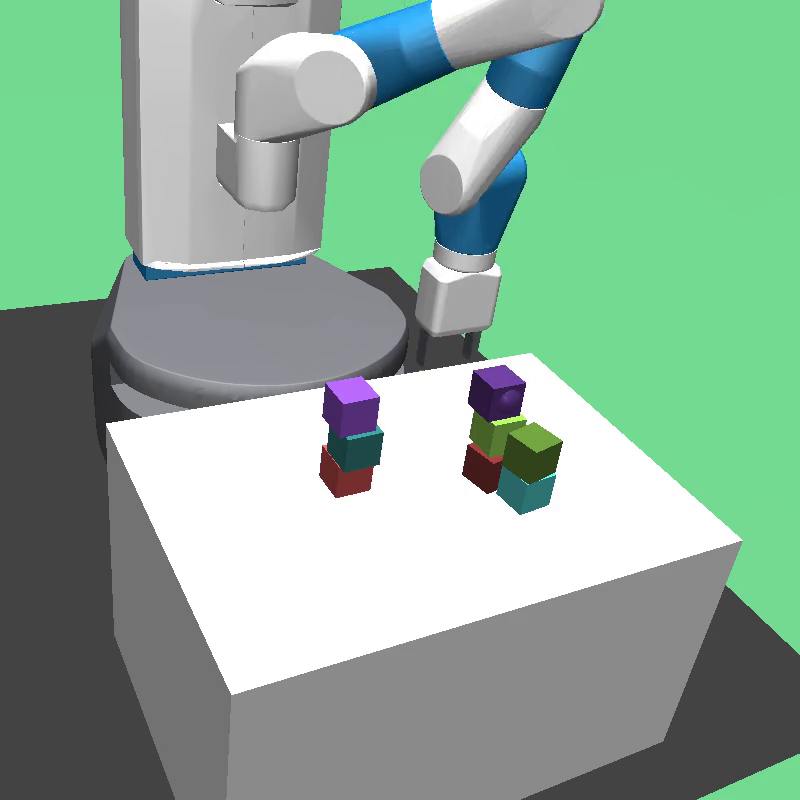}
\end{subfigure}
\begin{subfigure}[b]{0.31\linewidth}
    \centering
    \includegraphics[width=1.0\linewidth]{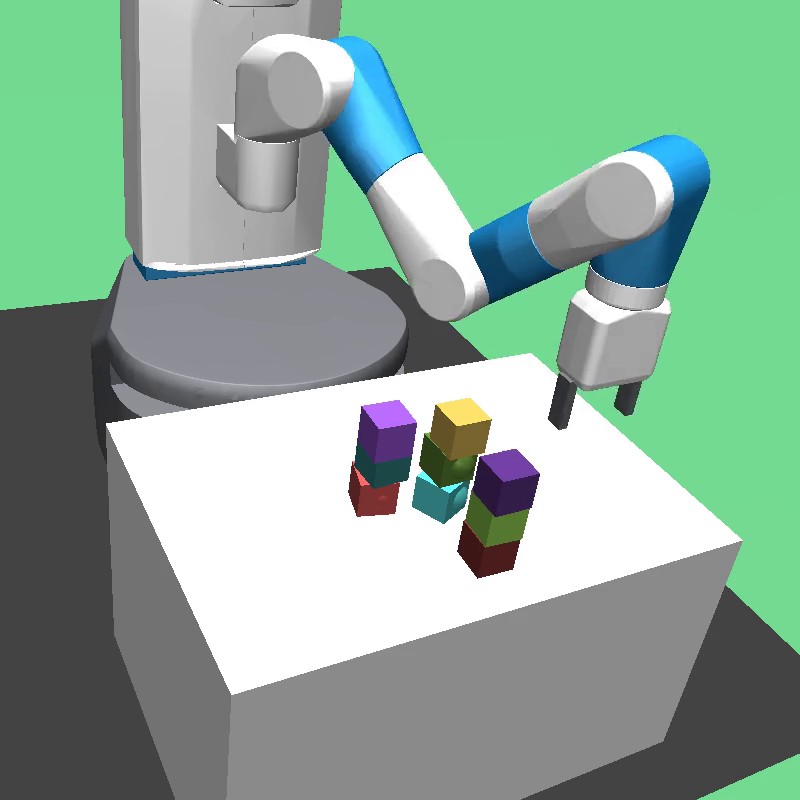}
\end{subfigure}
\caption{}
\end{subfigure}

\vskip 3pt
\caption{Visualizing goal configurations for (a) Pyramid 4 to Pyramid 7 and (b) Multiple Towers 4 to Multiple Towers 9.}
\label{fig:goal_configs}
\end{figure}

To visualize different goal types, we show frames from successful trajectories on Pyramid and Multiple Towers goal types in Figure \ref{fig:goal_configs}.

\subsection{Failure Modes}
\begin{figure}[h!]
\centering
\begin{tikzpicture}
  \node (img)  {\includegraphics[scale=.44]{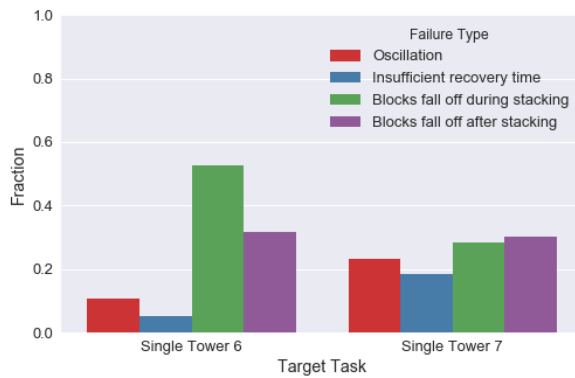}};
 \end{tikzpicture}
      \caption{Fraction of failures per failure type for a policy trained on Single Tower 6 and evaluated on target tasks Single Tower 6 and Single Tower 7. }  
    \label{fig:failures}
    \vspace{-4mm}
\label{fig:failure_types}
\end{figure}

We quantified the major failure modes of our system. They are as following:

\textbf{Oscillation:} The agent oscillates its end-effector without progressing towards the goal. Often, this happens when the target block is very close to the base of the tower. In this scenario, picking up the block risks toppling the tower. We hypothesize that in order to reduce this risk, the agent simply oscillates its end-effector. 

\textbf{Insufficient recovery time:} After 6 blocks have been stacked into a tower, the tower topples. The agent restarts stacking but is unable to stack all the blocks within the maximum time length of the episode. 

\textbf{Blocks fall off during stacking:} While stacking a tower of 6 blocks, the agent knocks one or more blocks off the table. Because the blocks are no longer on the table, the agent does not succeed. 

\textbf{Blocks fall off after stacking:} The agent succeeds in stacking a tower of 6 blocks, but the tower topples and block(s) fall off the table. 

These failures have been visualized in Figure~\ref{fig:failure_types} and the videos can be watched at: \url{https://richardrl.github.io/relational-rl}.

\subsection{Message Passing Rounds Ablation}
\begin{figure}[h!]
\centering
\begin{tikzpicture}
  \node (img)  {\includegraphics[scale=.33]{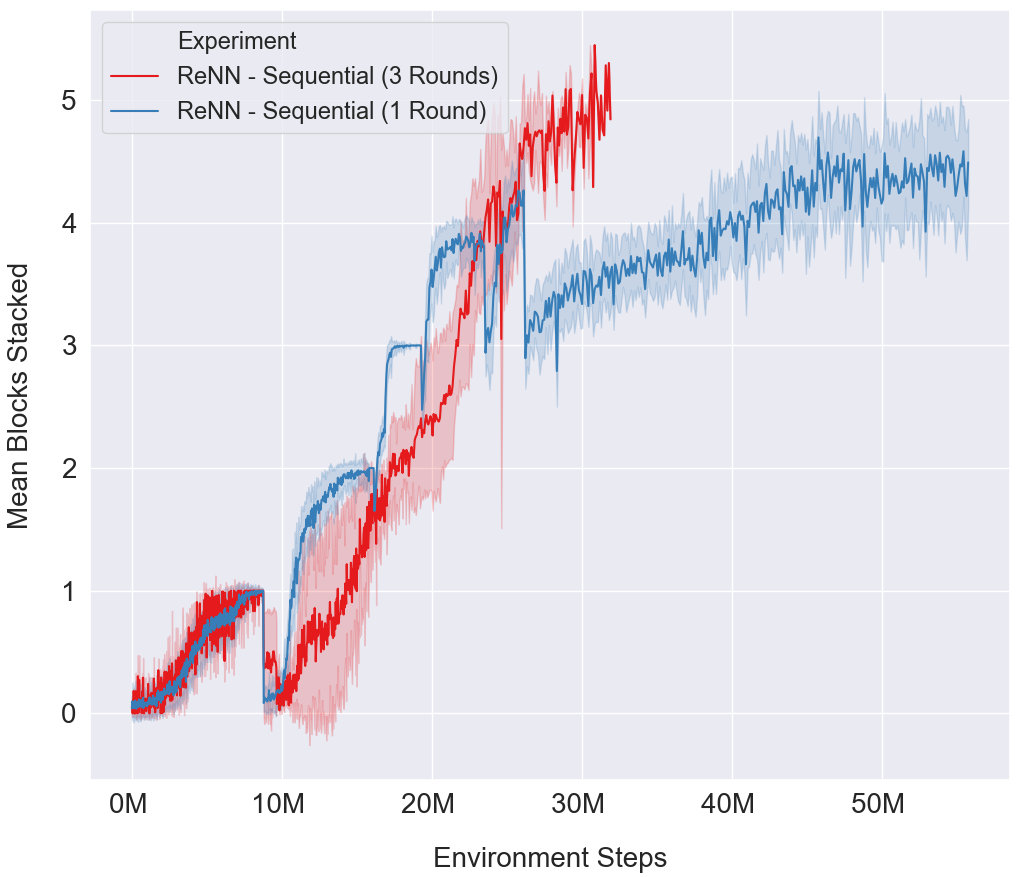}};
 \end{tikzpicture}
          \caption{Comparison of learning curves for \textit{ReNN - Sequential} with 3 and 1 messaging passing round(s) in the graph neural network.}
    \label{fig:failures}
    \vspace{-4mm}
\label{fig:mp_rounds_ablation}
\end{figure}

We compare the effect of different numbers of message passing rounds on convergence speed and accuracy. See Figure~\ref{fig:mp_rounds_ablation} and Table \ref{table:mp_rounds_esults}. A single message passing round converges faster in the earlier stages of the curriculum, but performs drastically worse when transferring the policy that learned to stack 5 blocks to stacking 6 blocks. It is likely that larger numbers of blocks require more rounds of message passing to accurately propagate information about multi-block stability.

\begin{table}[h!]
\caption{Success rates and convergence steps for one and three rounds of message passing in the graph neural network.}
\label{singletower_success_rates}
\begin{center}

\setlength\tabcolsep{4pt} 
\begin{tabular}{|c||c|c|c|c|}
\hline
Rounds & Single Tower 4 & Single Tower 5 & Single Tower 6\\\hline
1 & 90\% (23M)  & 77\% (26M) & 40\% (55M)\\
\hline
3 &\textbf{93\%}$\pm$4\% (\textbf{23M}) & \textbf{84\%}$\pm$6\% (\textbf{27M})&\textbf{75\%}$\pm$4\% (\textbf{30M})\\
\hline
\end{tabular}
\end{center}
\vspace{-4mm}
\label{table:mp_rounds_esults}
\end{table}
\end{appendix}

\end{document}